\documentclass[sigconf]{acmart}
\AtBeginDocument{%
  }

\setcopyright{none}
\usepackage{algorithm}
\usepackage{algorithmic}
\usepackage{multirow}
\usepackage{makecell}
\usepackage{tikz}
\usepackage{pgfplots}
\usepackage{soul}
\usepackage{xcolor}
\usepackage{listings}
\usepackage{colortbl}
\usepackage{rotating}

\def\shorten{\looseness=-1} 

\newcommand{\qald}{QALD-9}
\newcommand{\lcq}{LC-QuAD 1.0}

\newcommand{\kgqan}{\textsc{KGQAn}}

\newcommand{\sysName}{\textsc{Chatty-KG}}

\newcommand{\myNum}[1]{(\emph{#1})}

\definecolor{customcolor}{HTML}{C55A11}

\newcommand{\squishlist}{
  \begin{list}{$\bullet$}
   {
     \setlength{\itemsep}{0pt}
     \setlength{\parsep}{0pt}
     \setlength{\topsep}{0pt}
     \setlength{\partopsep}{0pt}
     \setlength{\leftmargin}{1.5em}
     \setlength{\labelwidth}{1em}
     \setlength{\labelsep}{0.5em} } }
\newcommand{\squishend}{
   \end{list}  }

\begin{document}



\title{{\sysName}: A Multi-Agent AI System for On-Demand Conversational Question Answering over Knowledge Graphs}


\author{Reham Omar}
\affiliation{%
  \institution{Concordia University}
  \country{Canada}
}

\author{Abdelghny Orogat}
\affiliation{%
  \institution{Concordia University}
  \country{Canada}
}

\author{Ibrahim Abdelaziz}
\affiliation{%
  \institution{IBM Research}
  \country{USA}
}

\author{Omij Mangukiya}
\affiliation{%
  \institution{Concordia University}
  \country{Canada}
}

\author{Panos Kalnis}
\affiliation{%
  \institution{KAUST}
  \country{Saudi Arabia}
}

\author{Essam Mansour}
\affiliation{%
  \institution{Concordia University} 
  \country{Canada}
}




\begin{abstract}
Conversational Question Answering over Knowledge Graphs (KGs) combines the factual grounding of KG-based QA with the interactive nature of dialogue systems. 
KGs are widely used in enterprise and domain applications to provide structured, evolving, and reliable knowledge. Large language models (LLMs) enable natural and context-aware conversations, but lack direct access to private and dynamic KGs. Retrieval-augmented generation (RAG) systems can retrieve graph content but often serialize structure, struggle with multi-turn context, and require heavy indexing. Traditional KGQA systems preserve structure but typically support only single-turn QA, incur high latency, and struggle with coreference and context tracking.
To address these limitations, we propose {\sysName}, a modular multi-agent system for conversational QA over KGs. {\sysName} combines RAG-style retrieval with structured execution by generating SPARQL queries through task-specialized LLM agents. These agents collaborate for contextual interpretation, dialogue tracking, entity and relation linking, and efficient query planning, enabling accurate and low-latency translation of natural questions into executable queries.\shorten
%
Experiments on large and diverse KGs show that {\sysName} significantly outperforms state-of-the-art baselines in both single-turn and multi-turn settings, achieving higher F1 and P@1 scores. Its modular design preserves dialogue coherence and supports evolving KGs without fine-tuning or preprocessing. Evaluations with commercial (e.g., GPT-4o, Gemini-2.0) and open-weight (e.g., Phi-4, Gemma 3) LLMs confirm broad compatibility and stable performance. Overall, {\sysName} unifies conversational flexibility with structured KG grounding, offering a scalable and extensible approach for reliable multi-turn KGQA.
\end{abstract}

\maketitle

\section{Introduction}
\label{sec:intro}

\begin{figure}[t]
  \centering
  \includegraphics[width=\linewidth, trim = .08cm 0.08cm .0cm .1cm, clip]{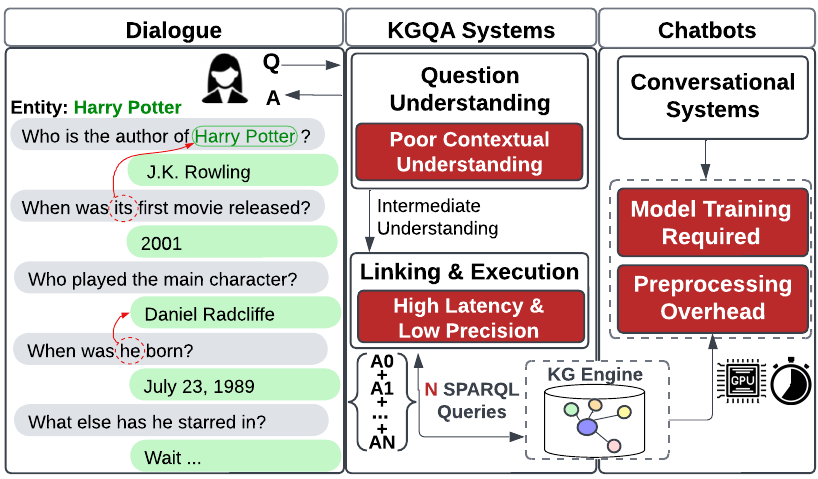}
  \vspace*{-4ex}
 \caption{Limitations of current KGQA and chatbot-based systems for real-time conversational access to arbitrary KGs. KGQA systems face issues with contextual understanding, query fragmentation, and latency. Chatbots offer better conversational QA over KGs but require expensive training and preprocessing, limiting adaptability and scalability. }
\label{fig:introduction}
\vspace*{-2ex}
\end{figure}

Conversational Question Answering over Knowledge Graphs (KGs) aims to combine the factual accuracy and structured reasoning of KG-based question answering with the natural, user-friendly experience of dialogue systems. KGs are widely adopted in domain-specific and enterprise applications to integrate heterogeneous data sources and provide reliable and up-to-date knowledge. Examples include general KGs, such as Wikidata~\footnote{\url{https://www.wikidata.org/}}, biomedical graphs like BioRDF~\footnote{\url{https://bio2rdf.org/}} and UMLS~\footnote{\url{https://www.nlm.nih.gov/research/umls}}
, academic KGs, such as DBLP~\footnote{\url{https://dblp.org/}}, and legal or financial graphs like FinDKG~\footnote{\url{https://xiaohui-victor-li.github.io/FinDKG/}}. Despite their strengths, interacting with KGs typically requires formal query languages, such as SPARQL or Cypher, creating a steep learning curve for non-expert users. This limits the deployment of conversational agents that can reliably answer questions over these knowledge sources.

To make KGs more accessible, Knowledge Graph Question Answering (KGQA) systems enable users to ask questions in natural language. However, most systems rely on domain-specific pipelines, require heavy KG preprocessing (e.g., EDGQA~\cite{EDGQA}), and depend on finetuned models (e.g., KGQAn~\cite{kgqan}), which limits their scalability and adaptability. These systems are typically designed for single-turn questions and often suffer from high latency. Such limitations make them unsuitable for multi-turn conversations that require context tracking and efficient processing. As shown in Figure~\ref{fig:introduction}, they also struggle with contextual understanding and multi-query coordination, resulting in low accuracy and slow response times, as demonstrated by KGQAn~\cite{kgqan} and EDGQA~\cite{EDGQA}.

Recent advancements in large language models (LLMs), such as GPT-4~\cite{gpt4o} and Gemini~\cite{geminiteam2024gemini15}, have greatly improved natural language understanding and reasoning. One might expect retrieval-augmented generation (RAG) to remove the need for explicit KG access by retrieving a subgraph and using it as context for the LLM. 
However, there is a gap between the RAG \emph{paradigm} and its current \emph{implementations}. 
In practice, graph-RAG systems such as Microsoft GraphRAG~\cite{graph_rag} must first interpret the question and extract entities before retrieval, but retrieval itself depends on heavy graph indexing and embedding stores that consume large memory and do not scale to large KGs. The retrieved subgraphs are then serialized into text, which discards structural information and weakens answer precision, especially for multi-entity or list-style queries. These systems may lack dialogue support, preventing multi-turn refinement or context reuse.
As illustrated in Figure~\ref{fig:introduction}, chatbot-based KG access (e.g., CONVINSE~\cite{convinse}, EXPLAIGNN~\cite{expalignn}) 
requires retraining or KG-specific preprocessing to remain accurate, making it costly and inflexible for evolving graphs. These limitations show that RAG-as-a-paradigm is promising, but current graph-RAG implementations fail to deliver accurate, structure-aware, and dialogue-capable KG access.\shorten

To address these limitations, we introduce {\sysName}, a modular multi-agent framework for conversational KGQA. The key technical novelty lies in combining RAG-as-a-paradigm with structured query execution through a training-free multi-agent design that requires no KG-specific preprocessing and generalizes across arbitrary graphs. Instead of asking an LLM to infer an answer from serialized triples, {\sysName} uses LLM agents to generate SPARQL queries over the KG, preserving graph semantics and reducing hallucinations. The pipeline is decomposed into lightweight agents for question understanding, dialogue context tracking, entity linking, and query generation, coordinated through a controller for low-latency execution. {\sysName} leverages recent LLM advances together with orchestration frameworks like LangGraph\footnote{\url{https://python.langchain.com/docs/langgraph/}} while avoiding costly retraining. This design enables real-time, verifiable access to \textbf{\textit{evolving KGs}} and supports both single-turn and multi-turn dialogues with minimal overhead.

\begin{table*}[t]
\vspace*{-2ex}
\small

\caption{Comparison of existing KGQA and conversational QA systems across key design aspects. KGQA systems lack multi-turn dialogue support and suffer from high latency, while conversational systems depend on fine-tuning and KG-specific preprocessing. {\sysName} overcomes these limitations with a LLM-powered multi-agent approach for real-time performance.
}
\vspace*{-2ex}
  \label{tab:back}
  \begin{tabular}{ccccccc}
    \toprule
     \textbf{System} & \textbf{Dialogue}&\textbf{Response Time}&\textbf{Question Understanding}&\textbf{Linking}&\textbf{Training}&\textbf{Pre-processing}\\
    \midrule
    \textbf{{\kgqan}}\cite{kgqan} & x & high& Fine-tuning& Semantic similarity based & \checkmark&x\\
    \textbf{{EDGQA}}\cite{EDGQA} & x & high&Constituency parsing& Index-based & x&\checkmark\\
    \midrule
    \textbf{CONVINSE\cite{convinse}} & \checkmark& low&Fine-tuning&Index-based& \checkmark&\checkmark\\
    \textbf{EXPLAIGNN\cite{expalignn}} & \checkmark & low&Fine-tuning&Index-based& \checkmark&\checkmark\\
    \midrule
    \textbf{{\sysName}} & \checkmark & low&LLM-Powered & LLM-Powered& x&x  \\
  \bottomrule
\end{tabular}
\end{table*}

Our extensive evaluation confirms that {\sysName} achieves outstanding performance across five real-world and large KGs of diverse application domains. It consistently outperforms state-of-the-art KGQA systems, including KGQAn~\cite{kgqan} for single-turn questions, 
RAG-based systems, such as GraphRAG~\cite{graph_rag} and ColBERT~\cite{santhanam2022colbertv2, khattab2020colbert},
and chatbot systems, such as GPT-4o, Gemini-2.0, CONVINSE~\cite{convinse} and EXPLAIGNN~\cite{expalignn} for multi-turn dialogue. {\sysName}'s modular design supports contextual reasoning and coherent dialogue without domain tuning. Experiments across a wide range of commercial (e.g., GPT-4o, Gemini-2.0) and open-weight LLMs (e.g., Phi-4, Gemma-3) demonstrate strong compatibility and stable performance. In general, {\sysName} offers a scalable, extensible, and reliable framework for conversational KGQA in real time with significant improvements in accuracy, efficiency, and adaptability.

In summary, our contributions are:
\begin{itemize}
     \item A modular multi-agent architecture for conversational KGQA, where individual agents can be swapped or enhanced independently, supporting future extensibility.
     \item Supports single- and multi-turn conversations with LLM-powered context tracking and disambiguation, while maintaining low latency for interactive use.
     \item Efficient SPARQL-based KG access via a query planning agent that eliminates the need for offline preprocessing.
     \item A comprehensive evaluation across five real KGs from diverse domains demonstrates {\sysName}'s superior performance in answer correctness and response time compared to state-of-the-art systems, with consistent results across commercial and open-weight LLMs.
\end{itemize}

\section{Background and Related Work}
This section surveys related work on KGQA, conversational models, and LLM-based agent frameworks. We outline the challenges of integrating conversational capabilities with structured knowledge and motivate our training-free and LLM-driven approach.


\subsection{Knowledge Graph Question Answering}
Conversational question answering over KGs typically requires three main steps: \myNum{i} Dialogue Understanding, which creates an abstract representation from the key entities and relations extracted from the question and conversation history; \myNum{ii} Linking, which maps the abstract representation to vertices and predicates from the KG; and \myNum{iii} Answer Generation, which uses the abstract representation and the mapped KG vertices and predicates to either extract the answer from the KG or generate it using a trained model. In this section, we discuss the two categories of systems summarized in \autoref{tab:back}: KGQA; e.g., \cite{kgqan,EDGQA}, and Conversational Systems; e.g., \cite{convinse,expalignn} and how it compares to our approach.

\textbf{KGQA systems} rely on generating SPARQL queries that represent an individual question. Examples include  gAnswer~\cite{gAnswer2014, gAnswer2018}, NSQA~\cite{NSQA}, and WDAqua~\cite{WDAquaWWW18}. The top two performing systems, KGQAn~\cite{kgqan} and EDGQA~\cite{EDGQA}, are examined in greater detail in this section to better understand their design choices, strengths, and limitations. KGQAn is the state-of-the-art across multiple datasets \cite{kgqan}, including {\qald} \cite{qald9}, while EDGQA achieves strong results on the {\lcq} dataset \cite{EDGQA,lcquad}. However, these systems cannot handle conversational interactions as they are designed for single-turn and standalone questions. KGQAn's pretrained models are limited by their training on single-turn queries and a relatively small context window length that cannot fit conversation history. Similarly, EDGQA's rule-based approach does not scale to multi-turn conversations.
Both KGQAn and EDGQA have high response times, making them unfit for conversational use. KGQAn's latency is due to repeated similarity-server requests, numerous SPARQL queries execution for high recall, and an answer-type filtering step. EDGQA also runs multiple SPARQL queries and uses three different systems for linking, contributing to its delay~\cite{kgqan}.

For question understanding, KGQAn fine-tunes an LLM to generate triples from questions, requiring a manually created dataset that is time-consuming and subject to annotator bias. It also trains a separate answer type detection model. In contrast, EDGQA uses the Stanford CoreNLP parser with rules tailored to a specific dataset, limiting cross-domain applicability. For linking, KGQAn makes hundreds of similarity-server calls, slowing response time, while EDGQA uses three indexing systems (Falcon~\cite{falcon}, EARL~\cite{earl}, Dexter~\cite{dexter}), requiring heavy pre-processing and reducing adaptability to new knowledge graphs. Overall, KGQAn demands extensive model training, while EDGQA relies on significant pre-processing.

\begin{figure*}[t]
\vspace*{-2ex}
  \centering
\includegraphics[width=0.98\linewidth]{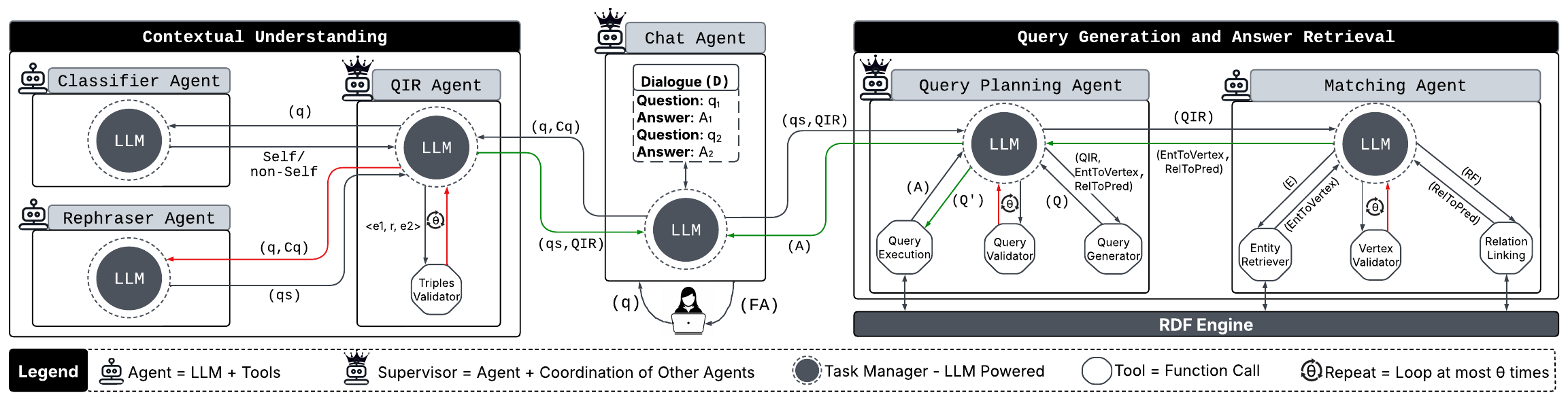}
\vspace*{-3ex}
\caption{\sysName{}'s hierarchical multi-agent architecture. A top-level \textit{Chat Agent} coordinates two supervised modules: \textit{Contextual Understanding} and \textit{Query Generation \& Answer Retrieval}, each composed of specialized LLM-powered agents. This design enables modular, low-latency KGQA without training or pre-processing, and improves adaptability and real-time performance.\shorten}
\label{fig:framework-agent}
\vspace*{-2ex}
\end{figure*}

\textbf{Conversational systems} rely on information retrieval techniques to extract or generate answers. CONVINSE~\cite{convinse} and EXPLAIGNN~\cite{expalignn} are end-to-end conversational question-answering systems on heterogeneous data sources, including KGs, text, and tables. These systems, which only support the Wikidata KG \cite{wikidata},  handle conversational interactions with low response times, making them suitable for real-time user interaction. For question understanding, they transform questions into intent-explicit structured representations (SR) using fine-tuned transformer models \cite{transformer} (BART \cite{bart} and T5 \cite{t5}). 
CONVINSE and EXPLAIGNN formalize linking as evidence retrieval and scoring. They use Clocq \cite{clocq} for named entity disambiguation and KG-item retrieval. Clocq requires KG pre-processing before KG facts retrieval. For answer generation, CONVINSE trains a Fusion-in-Decoder (FiD) model \cite{fid} to generate answers based on top-ranked evidence. EXPLAIGNN constructs a heterogeneous answering graph from retrieved entities and evidence. 
Both systems require model training for question understanding and answer generation. They also require KG pre-processing, which limits their adaptability to new KGs.

In summary, existing systems face key limitations that hinder effective conversational question answering over diverse and evolving KGs. Traditional KGQA systems lack support for multi-turn dialogue and suffer from high latency. Conversely, conversational systems improve interactivity and speed but depend heavily on fine-tuning and KG-specific pre-processing, reducing flexibility across domains. In contrast, {\sysName} introduces a modular, low-latency conversational framework that eliminates the need for training or pre-processing. It leverages LLMs for dynamic dialogue tracking, question understanding, and entity linking to enable scalable and adaptable QA over arbitrary knowledge graphs.

\subsection{LLM Prompting Strategies}
LLMs such as GPT-4, Gemini, and Qwen~\cite{qwen2.5} have shown strong language understanding and generation skills across diverse tasks. 
Their generalization without task-specific training makes them especially valuable in modular systems like {\sysName}, where they serve as interchangeable agents for reasoning, classification, or generation. Prompt engineering has emerged as a lightweight, cost-effective alternative to fine-tuning by steering LLMs through carefully crafted instructions~\cite{prompt_spec}. Common prompting strategies include: 1) zero-shot prompting with only a task description is provided to the LLM, 2) few-shot prompting which adds to the LLM a handful of examples , and 3) chain-of-thought (CoT) prompting, which guides intermediate reasoning steps before producing an answer~\cite{cot, zeroshot_cot}. The best strategy depends on the task: classification or retrieval often works well with zero-shot prompts, while generation or complex reasoning benefits from few-shot or CoT prompting.
In {\sysName}, simpler agents (e.g., for classification or routing) use zero-shot prompts for speed, while agents handling contextual understanding or answer synthesis apply few-shot or CoT prompting to support multi-step reasoning. This flexibility optimizes both efficiency and quality across the multi-agent pipeline.

\subsection{LLM Agents}
LLM agents are emerging systems that use the reasoning, planning, and language capabilities of large language models to autonomously complete tasks. At their core, these agents treat the LLM as a central decision-maker that interacts with tools, APIs, and environments in a loop of observation, reasoning, and action. Recent frameworks, such as ReAct~\cite{yao2023react}, AutoGPT~\cite{gravitasauto}, and BabyAGI~\cite{nakajima2023task}, combine LLMs with tool use and planning mechanisms. More structured frameworks like LangChain, LlamaIndex\footnote{\url{https://www.llamaindex.ai/}}, and OpenAgents~\cite{OpenAgents} provide abstractions for integrating tools, retrievers, memory, and control flows. LangChain enables modular pipelines that compose LLMs with APIs, databases, and reasoning components. LangGraph\footnote{\url{https://python.langchain.com/docs/langgraph/}}, a recent LangChain extension, introduces stateful graph-based control for defining multi-agent workflows with dynamic routing.

A key enabler of these systems is tool calling (also known as function calling), where the LLM invokes external tools at inference time to offload specific tasks, such as computation, API queries, or structured data access. Tool calling has become central to modern agent frameworks, allowing models to extend beyond generation and act as orchestrators. In our system, agents selectively call tools based on their roles, for example, to resolve entity mentions, fetch KG subgraphs, or execute SPARQL queries. This targeted, task-specific integration enables agents to remain lightweight and model-agnostic, while grounding their decisions in accurate, up-to-date KG content.\shorten

LLM agents are increasingly used in domains, such as web automation~\cite{yang2024agentoccam}, code generation~\cite{wang2024executable}, and data wrangling~\cite{li2024towards}. In these settings, they break down complex goals, retrieve intermediate results, and adapt to user feedback~\cite{wang2024survey}. However, most systems are tailored to unstructured or semi-structured data and struggle with formal sources, such as relational databases or knowledge graphs (KGs)~\cite{shi2024survey}. Challenges in schema alignment, entity linking, and precise query generation remain underexplored. Therefore, the use of LLM agents for KG question answering (KGQA) is still emerging. In this work, we demonstrate how core agent capabilities, such as contextual understanding, linking, and modular reasoning, can be applied to KGQA in a lightweight and efficient manner. Our system avoids complex planning or orchestration. Instead, our agents use prompt-guided LLM calls to interpret questions, link KG elements, and generate a minimal set of executable queries.

\section{{\sysName} Architecture}


{\sysName} adopts a hierarchical design that splits the task into two main subtasks: language-level processing and graph-level reasoning. Language-level tasks include question classification, ambiguity resolution, and intermediate representation generation. Graph-level tasks cover entity/relation linking and SPARQL query construction. Each subtask is handled by a cluster of LLM agents. Each agent wraps a \emph{task-managed LLM}, a language model guided by role-specific logic and tools. Some agents act as supervisors, delegating to subordinate agents. 
%
This modular and domain-agnostic architecture supports accurate reasoning, flexible operation across KGs, and efficient execution without retraining or KG-specific preprocessing. The matching agent does not assume a fixed schema and instead dynamically extracts relevant subgraphs using SPARQL queries. Key stages incorporate validation and retry mechanisms to mitigate error propagation. We first define the main representations.

\begin{definition}[Dialogue]
\vspace*{-0.9ex}
The dialogue history is defined as $\mathcal{D} = \{(q_1, A_1), (q_2, A_2), \dots, (q_{n-1}, A_{n-1}) \}$, where each $q_i$ is a user-issued question and $A_i = [a_1, a_2, \dots, a_m]$ is the corresponding system-generated answer. Each $a_j$ may be a count, a Boolean value, or an entity; $A_i$ may also be a list of such values.
\vspace*{-.9ex}
\end{definition}

\begin{definition}[Question Intermediate Representation (QIR)]
Given a question \( q \) that mentions entities, expresses relations, and targets an unknown variable, potentially involving intermediate variables for multi-hop reasoning, the \textit{QIR} is defined as \( QIR = (E, U, R, RF) \). Here, \( E \) is the set of mentioned entities, \( U \) is the set of variables (including the target and any intermediates), $\mathcal{R}$ is the set of relation phrases in $q$, and \( RF \) is a set of relational facts. Each \( rf \in RF \) is a triple \( \langle e_1, r, e_2 \rangle \), where \( e_1, e_2 \in (E \cup U) \) and \( r \in \mathcal{R}\).
\vspace*{-0.5ex}
\end{definition}

{\sysName} is structured into three core modules: \textit{Contextual Understanding}, \textit{Chat Management}, and \textit{Query Generation and Answer Retrieval} (\autoref{fig:framework-agent}). Each module contains one or more agents and tools with well-defined communication interfaces. The system operates on a dialogue $\mathcal{D}$, using its history to form the context $C_q$ for interpreting new user questions. Upon receiving a new question $q_i$, the \textbf{Chat Agent} initiates processing by forwarding $q_i$ and $C_q$ to the \textbf{QIR Agent}, which builds the intermediate representation to guide query generation.
To manage complexity, the QIR Agent delegates two subtasks to specialized agents. First, the \textbf{Classifier Agent} determines whether $q_i$ is \emph{\ul{self-contained}} (interpretable alone) or \emph{\ul{context-dependent}} (requiring prior turns). For example, “Who is the author of Harry Potter?” is self-contained, while “When was its first movie released?” depends on earlier context.\shorten

The classification ensures downstream agents receive clear and unambiguous input. If the question is classified as context-dependent, the QIR Agent forwards both $q_i$ and $C_q$ to the \textbf{Rephraser Agent}. This agent reformulates the question into a self-contained version using dialogue context. For example, it transforms “When was its first movie released?” into “When was the first Harry Potter movie released?”. The QIR Agent then processes the resulting question, whether rephrased or original, to generate the Question Intermediate Representation (QIR). This involves extracting a set of triples representing the question’s semantics.

{\sysName} uses an assertion-based validation approach to ensure output correctness. Inspired by software assertions, it defines task-specific structural and semantic checks that are independent of input content. For example, the \textbf{Triples Validator} checks that each extracted triple has valid components (subject, predicate, object).
If validation fails, the QIR Agent retries generation up to a threshold $\theta$. Once validated, the triples form the QIR. For instance, the QIR for the rephrased question is \{\textit{E: \{"Harry Potter"\}, U: \{"?year"\}, RF: \{$\langle$Harry Potter, released, ?year$\rangle$\}}\}.\shorten

The \textbf{Query Planning Agent} receives the QIR and identifies a minimal set of SPARQL queries to retrieve the answer. It begins by delegating entity and relation linking to the \textbf{Matching Agent}, which uses the \textbf{Entity Retriever} and \textbf{Relation Linking} tools to map QIR elements to KG URIs. The resulting mappings are passed to the \textbf{Vertex Validator}, which checks that each entity is syntactically correct and appears in the candidate list. If validation fails, the system retries up to $\theta$ times. After successful validation, the URIs are returned to the planning agent for query construction.

The \textbf{Query Generator} constructs candidate SPARQL queries. The \textbf{Query Planning Agent} selects a minimal subset for execution, typically one to three queries, based on its understanding, without relying on a fixed threshold. The selected queries are passed to the \textbf{Query Validator}, which uses assertions to ensure syntactic and semantic correctness, retrying on failure.
This avoids executing irrelevant queries and reduces overhead. The selected queries are sent to the \textbf{Query Execution} module to retrieve the raw answer $A$. The Chat Agent may rephrase it into a user-friendly response $FA$, e.g., turning \texttt{"2001"} into “The first Harry Potter movie was released in 2001.” The final answer is returned to the user, and the dialogue $\mathcal{D}$ is updated to complete the interaction.

\section{Chat Agent and Contextual Understanding}
\label{sec:dialogue}

This section introduces two core components: the \textit{Chat Agent}, which manages dialogue state and controls workflow, and the \textit{Contextual Understanding} module, which transforms natural language questions into a structured Question Intermediate Representation.

\subsection{Chat Agent}
The \textbf{Chat Agent} acts as the top-level controller in {\sysName}'s multi-agent framework. It is responsible for: \myNum{i} orchestrating module execution, \myNum{ii} maintaining dialogue history, and 
\myNum{iii} reformulating answers and enabling multilingual queries through LLM-based translation.\shorten
%
When a new user question \( q_n \) arrives, the Chat Agent retrieves the dialogue history $\mathcal{D}$ and constructs the corresponding context $C_q$ needed to interpret the question. Since $C_q$ is passed as input to prompt-based LLM agents (e.g., the Rephraser Agent), it must remain within the LLM's token limit. We define \( C_q = \{(q_i, A_i^L)\}_{i=1}^{n-1} \), where each \( A_i^L \) is a truncated prefix
of \( A_i \), limited to \( L \) items when \( |A_i| > L \). This preserves essential context while ensuring compact prompts.
The pair \( (q_n, C_q) \) is passed to the Contextual Understanding module, which generates the structured Question Intermediate Representation (QIR). The QIR is then forwarded to the Query Generation and Answer Retrieval module, which constructs and executes the corresponding SPARQL query and returns the raw answer \( A_n \).
Optionally, the Chat Agent reformulates \( A_n \) into a more natural final answer \( FA_n \) using a zero-shot prompting strategy:

\begin{equation}     
    \label{eq:reformulate-final-answer}     
    FA_n = f(A_n, q_n, I_{FA}),
\end{equation}

where \( I_{FA} \)\footnote{All prompts are available in the \href{https://github.com/CoDS-GCS/Chatty-KG/blob/main/Appendix.pdf}{\color{blue}\ul{supplementary materials.}}} is a prompt instructing the LLM to express short literal answers in fluent natural language. We adopt zero-shot prompting due to the simplicity and consistency of the task, which typically involves converting short outputs into complete sentences.
Finally, the Chat Agent appends \( (q_n, A_n) \) to the dialogue history and awaits the next question. This process enables context-aware, multi-turn interaction while maintaining low latency and prompt efficiency.\shorten

\begin{algorithm}[t]
\caption{Contextual Understanding Pipeline}
\label{alg:dialogue_processing}
\begin{flushleft}
\textbf{Input:} \( q_i \): User question, \( C_q \): Question context, \( \theta \): Retry threshold, \( I_C \): Classification instructions, \( I_{RF} \): Reformulation instructions, \( I_{QU} \): Understanding instructions\\
\textbf{Output:} $QIR = (E, U, R, RF)$: Intermediate Representation
\end{flushleft}
\begin{algorithmic}[1]
\STATE \( q_{\text{type}} \gets \text{ClassifierAgent}(q_i, I_C) \)
\label{algo1:line:q_type}
\IF{\( q_{\text{type}} = \text{`Dependent'} \)}
    \STATE \( q_{is} \gets \text{RephraserAgent}(q_i, C_q, I_{RF}) \)
    \label{algo1:line:q_s_agent}
\ELSE
    \STATE \( q_{is} \gets q_i \)
\ENDIF
\STATE \( \text{valid} \gets \text{False} \)
\WHILE{\( \text{valid} = \text{False} \land \theta > 0 \)}
    \STATE \( triples \gets \text{QIRAgent.GenerateTriples}(q_{is}, I_{QU}) \)
    \label{algo1:line:triples}
    \STATE \( \text{valid} \gets \text{QIRAgent.ValidateTriples}(triples) \)
    \label{algo1:line:valid}
    \STATE \( \theta \gets \theta - 1 \)
\ENDWHILE
\STATE \( QIR \gets \text{QIRAgent.ConstructQIR}(triples)\)
\STATE \textbf{return} \( QIR \)
\end{algorithmic}
\end{algorithm}

\subsection{Contextual Understanding}

To interpret a user question \( q_i \) in its dialogue context \( C_q \), the Contextual Understanding module runs a multi-stage pipeline that converts natural language into a structured Question Intermediate Representation (QIR). This process is managed by the \textit{QIR Agent}, which coordinates two key subtasks handled by auxiliary agents: classification and reformulation.
As shown in Algorithm~\ref{alg:dialogue_processing}, the \textit{Classifier Agent} first determines whether \( q_i \) is self-contained or context-dependent. If the question depends on prior context, the \textit{Rephraser Agent} generates a standalone version \( q_{is} \) using \( C_q \) to resolve references and omissions. Otherwise, \( q_{is} \leftarrow q_i \).

The unified question \( q_{is} \) is then processed by the QIR Agent to extract semantic triples via prompt-guided LLM interaction. These triples are validated for structure and meaning, and the agent retries up to \( \theta \) times if validation fails. Once validated, the resulting QIR is returned to the Chat Agent for downstream processing.
Separating these subtasks improves LLM performance by reducing prompt complexity and avoiding instruction overload, as each agent uses a dedicated session and objective-specific prompt.

\subsubsection{Classifier Agent}

The Classifier Agent identifies whether a question is \emph{self-contained} or \emph{context-dependent} by detecting unresolved references or pronouns that imply reliance on earlier dialogue turns. It uses a zero-shot prompting strategy with a natural-language instruction \( I_C \) that explains the classification criteria and guides the decision.
We avoid using traditional machine learning classifiers, which require labeled training data that is costly and often unavailable. Moreover, this task goes beyond feature-level classification such as pronoun presence. For instance, “When was the first movie released?” appears complete but is context-dependent if the subject entity was mentioned earlier. The LLM-based approach enables semantic reasoning over such implicit dependencies.
The agent returns a label \( q_{\text{type}} \in \{\text{Self-contained}, \text{Dependent}\} \), which determines the subsequent processing path. The classification step is formalized as:
\begin{equation}
    \label{eq:classify}
    q_{\text{type}} = f(q_i, I_C)
\end{equation}

\subsubsection{Rephraser Agent}
The Rephraser Agent converts a context-dependent question \( q_i \) into a self-contained version \( q_{is} \) by resolving implicit references using the dialogue history \( C_q \). This is achieved through a prompt-driven LLM function:

\begin{equation}
    \label{eq:reformulate}
    q_{is} = f(q_i, C_q, I_{RF})
\end{equation}

Here, the model receives the question \( q_i \), its context \( C_q \), and an instruction prompt \( I_{RF} \) that specifies the criteria for semantic completeness. The prompt directs the model to resolve dependencies—such as pronouns, ellipses, or anaphora—and rewrite the question as a standalone utterance that no longer depends on prior turns.
Unlike rule-based systems that rely on handcrafted heuristics, this approach leverages the LLM’s pretrained reasoning to generalize across varied linguistic forms. Although the task involves text generation, it is bounded in scope: the objective is to restructure the input without introducing new content. This constraint makes it well-suited for zero-shot prompting. The resulting question \( q_{is} \) preserves the original intent of \( q_i \) while satisfying structural and semantic requirements for QIR generation.


Ambiguous or conflicting entity mentions across dialogue turns are resolved by the rephraser agent using the full dialogue history. The history stores explicit entity mentions in \(C_q\), allowing the agent to replace pronouns with the correct referenced entity before reformulating the question. Since the agent relies on LLMs with strong long-context coreference abilities, it can disambiguate pronouns and produce a clear, standalone version of the user query.

\subsubsection{QIR Agent}
The QIR Agent converts the self-contained question \( q_{is} \) into a structured semantic form. It generates a set of triples \( RF = \{ \langle s_i, r_i, o_i \rangle \}_{i=1}^k \), where \( s_i, o_i \in E \cup U \) and each relation \( r_i \in \mathcal{R} \), the set of relation phrases in $q_{is}$. These triples represent the core semantic structure of the question and are incorporated into the final QIR only after passing a validation step.
An initial prompting method guided the LLM to extract triples by clarifying variable-entity distinctions, highlighting relevant facts, and enforcing output format. However, this led to frequent issues such as omitted variables and generic or ambiguous predicates (e.g., \textit{has}) that distorted intent.
To address this, the QIR Agent adopts a chain-of-thought (CoT) prompting strategy combined with few-shot learning. The CoT prompt decomposes the task into three reasoning steps: \myNum{i} identifying key entities, \myNum{ii} detecting unknowns (answers or intermediate variables), and \myNum{iii} generating verb- or noun-based relations between them. This structured reasoning improves precision and mitigates earlier shortcomings. Two examples (one with verb-based, one with noun-based relations) are included to guide generation.
The agent receives a declarative prompt \( I_{QU} \) and example set \( e \), returning a structured output in JSON format for downstream use. The process is formalized as:
\begin{equation}
\label{eq:understanding}
RF = f(q_{is}, I_{QU}, e)
\end{equation}


\noindent \textbf{Triples Validator.} After generating the candidate triples, the Triples Validator checks the output against structural and semantic constraints. Specifically, the result must satisfy: (1) it is a well-formed JSON object, (2) each triple includes a valid subject, predicate, and object, (3) at least one triple references a known entity
, and (4) for non-boolean questions, at least one variable 
must appear. Boolean questions such as ``Is Michelle the wife of Barack Obama?'' are exempt from the fourth condition, as they may consist entirely of grounded triples like \( \langle \text{Michelle}, \textit{wife of}, \text{Barack Obama} \rangle \). If the output fails validation, the QIR Agent retries generation up to a threshold \( \theta \). Once validated, the QIR is constructed from the validated triples and returned.

\subsubsection{Time and Space Complexity}
Algorithm~\ref{alg:dialogue_processing} executes a bounded sequence of LLM agent calls, whose cost is dominated by language model interactions. Let \(C_{\text{LLM}}\) denote the cost of one LLM invocation. The algorithm performs at most three LLM calls: one to the Classifier Agent (Line~\ref{algo1:line:q_type}), one conditional call to the Rephraser (Line~\ref{algo1:line:q_s_agent}), and up to \( \theta \) calls to the QIR Agent for triple generation (Line~\ref{algo1:line:triples}). Thus, the total time complexity is \(O(\theta \cdot C_{\text{LLM}})\), since the number of LLM calls is constant. The validation step (Line~\ref{algo1:line:valid}) uses lightweight rule-based checks and runs in constant time. Space complexity is dominated by the dialogue history \( \mathcal{D} \), which adds one QA pair per turn, resulting in \(O(n)\) space for \(n\) turns.

\section{Query Generation and Answer Retrieval}
\label{sec:query_agent}


This section describes the \textit{Query Planning Agent}, which converts the QIR into a small set of SPARQL queries to retrieve the answer \(A\). Entity and relation linking are delegated to the \textit{Matching Agent}. The agent executes only a high-confidence subset of candidate queries, requiring no KG-specific preprocessing, remaining compatible with standard RDF backends, and reducing latency.

\subsection{Matching Agent}

Given the \( QIR = (E, U, R, RF) \), the Matching Agent aligns QIR entities and relations to their corresponding URIs in the target KG by querying its SPARQL endpoint at runtime. As shown in Algorithm~\ref{alg:vertex_linking}, for each entity \( e \in E \) (Line \ref{algo2:line:for_1}), the \textit{Entity Retriever} issues a keyword-based SPARQL query to retrieve up to \( v_{limit} \) candidate vertices (Line \ref{algo2:line:v_list}). Vertex selection is then performed via prompt-guided reasoning using instructions \( I_{VL} \) (Line \ref{algo2:line:v_chosen}). The result is validated by the \textit{Vertex Validator}, which checks both membership in the candidate set and JSON correctness (Line \ref{algo2:line:v_valid}). If validation fails, selection is retried up to \( \theta \) times (Lines \ref{algo2:line:try_start}–\ref{algo2:line:try_end}) to mitigate LLM hallucinations or formatting errors. Validated mappings are stored in \( EntToVertex \) (Line \ref{algo2:line:EntToVertex}).
Next, for each relation \( r \in R \) (Line \ref{algo2:line:Rf-for_start}), the \textit{Relation Linking Tool} retrieves and ranks candidate predicates to maximize recall. The resulting ranked list is stored in \( RelToPred \) (Line \ref{algo2:line:RelToPred}). Finally, the agent returns: (i) \( EntToVertex: E \rightarrow V \), mapping each entity to a KG vertex, and (ii) \( RelToPred: R \rightarrow \text{List}(P) \), mapping each relation to a list of semantically relevant predicates\footnote{Here, $\text{List}(P)$ denotes a list of predicates, allowing each relation to link to multiple KG predicates rather than a single one.}.
This runtime-only, modular design avoids KG-specific preprocessing, schema alignment, or offline indexing. As a result, it generalizes well across diverse KGs while maintaining high precision and adaptability.

\subsubsection{Entity Retriever}

Entity linking aims to match each known entity \( e \in E \) in the \( QIR = (E, U, R, RF) \) to a unique, semantically appropriate vertex \( v \in V \) in the Knowledge Graph (KG). Formally, this defines a mapping \( f: E \rightarrow V \), where each entity is aligned to one KG vertex. To ensure compatibility with arbitrary SPARQL endpoints and avoid KG-specific preprocessing or offline indexing, the Entity Retriever dynamically queries the RDF engine using keyword-based SPARQL. It extracts candidate vertices whose labels contain words from the entity name.
The number of candidates per entity is limited by a configurable parameter \( v_{\text{limit}} \), 
Once the list \( v_{\text{list}} \subseteq V \) is retrieved for an entity \( e \), the Matching Agent invokes a prompt-guided LLM to select the best candidate:

\begin{equation}
    \label{eq:vertex}
    v = f(e, v_{\text{list}}, I_{VL}),
\end{equation}

where \( I_{VL} \) is a handcrafted prompt instructing the LLM to assess label similarity, favoring exact matches when available. To support weaker models, the prompt includes a definition of "exact match." Only vertex labels (not URIs) are provided to improve interpretability, and the output is formatted in JSON for reliable parsing.
This zero-shot classification setup allows the LLM to act as a semantic filter under in-context learning, enabling generalization across varied and domain-specific KGs. By decoupling retrieval from selection and avoiding retraining, the Entity Retriever ensures modularity, extensibility, and adaptability without manual schema alignment.

\begin{algorithm}[t]
\caption{Vertex and Relation Linking}
\label{alg:vertex_linking}
\begin{flushleft}
\textbf{Input:} $QIR = (E, U, R, RF)$: Intermediate representation, $v_{limit}$: Maximum number of retrieved vertices, $endpoint$: SPARQL endpoint, $I_{VL}$: Vertex linking prompt, $\theta$: Retry threshold\\
\textbf{Output:} $EntToVertex$: Entity to linked vertices map, $RelToPred$: Relation to linked predicates map
\end{flushleft}
\begin{algorithmic}[1]
\STATE $EntToVertex, RelToPred \gets \{\}, \{\}$
\FOR{each $e \in QIR.E$}
\label{algo2:line:for_1}
    \STATE $v_{list} \gets \text{MAgent.entity\_retriever}(e, v_{limit}, endpoint)$
    \label{algo2:line:v_list}
    \STATE $valid \gets \text{False}$
    \WHILE{$valid = \text{False} \land \theta > 0$}
    \label{algo2:line:try_start}
        \STATE $v_{chosen} \gets \text{MAgent}(e, v_{list}, I_{VL})$
        \label{algo2:line:v_chosen}
        \STATE $valid \gets \text{MAgent.Validate}(v_{chosen}, v_{list})$
        \label{algo2:line:v_valid}
        \STATE $\theta \gets \theta - 1$
    \ENDWHILE
    \label{algo2:line:try_end}
    \STATE $EntToVertex[e] \gets v_{chosen}$
    \label{algo2:line:EntToVertex}
\ENDFOR
\FOR{each $rf \in QIR.RF$}
\label{algo2:line:Rf-for_start}
    \STATE $preds \gets \text{MAgent.rel\_Linking}(rf, endpoint, EntToVertex)$
    \label{algo2:line:preds}
    \STATE $RelToPred[rf.r] \gets preds$
    \label{algo2:line:RelToPred}
\ENDFOR
\STATE $\textbf{Return EntToVertex, RelToPred}$
\end{algorithmic}
\end{algorithm}

\subsubsection{Vertex Validator}

The Vertex Validator ensures semantic and syntactic correctness of the selected vertex \( v \) before finalizing entity alignment. It enforces two key constraints. First, it verifies that the LLM’s output is a syntactically valid and parsable JSON object, suitable for downstream processing. Second, it checks that the selected vertex \( v \) belongs to the retrieved candidate list \( v_{\text{list}} \), guaranteeing that the output corresponds to a real KG entity and not a hallucinated or out-of-scope result.  If either validation step fails, the Matching Agent re-invokes the vertex selection tool with a retry budget bounded by the threshold \( \theta \). This mechanism ensures that only grounded, well-formed vertex selections are propagated to subsequent stages, thereby preserving system robustness and preventing semantic drift in the query generation process.

\subsubsection{Relation Linking}

For relation linking, assigning a single predicate to each semantic relation is often insufficient, as entities in the KG may be connected through multiple semantically related predicates. For example, in DBpedia, the entity \textit{Intel} is linked via predicates such as \texttt{founders}, \texttt{founder}, and \texttt{foundedBy}. To accurately answer a question like “Who founded Intel?”, the system must consider all such alternatives to ensure comprehensive coverage.
The Relation Linking tool addresses this by associating each relation \( r \in QIR.R \) with a ranked list of candidate predicates from the KG. For a given relation \( r \), the tool queries the SPARQL endpoint to retrieve all predicates that connect the previously linked source and target entities. Each predicate URI is converted into a human-readable label by extracting its final segment. Both the relation label and predicate labels are encoded using BERT~\cite{bert}, and their semantic similarity is measured via cosine similarity. The resulting predicates are ranked and stored in the \( RelToPred \) mapping, allowing the downstream query planner to select the most appropriate predicates without relying on KG-specific rules or preprocessing.

\subsubsection{Time and Space Complexity}

The time complexity of the vertex and relation linking phase is \(O(q_{\text{cost}} + \theta \cdot C_{\text{LLM}})\), where \(q_{\text{cost}}\) is the cost of SPARQL queries used to retrieve candidate vertices and predicates. Since each QIR has only a few entities and relations (typically <10), both SPARQL access and LLM inference operate on bounded input sizes, and validation runs in constant time. The value of \(q_{\text{cost}}\) is implementation-dependent and varies with the RDF engine. For space complexity, storing one vertex per entity yields constant overhead for \(\texttt{EntToVertex}\), while the main cost comes from \(\texttt{RelToPred}\), which stores ranked predicates. Let \(p_{\text{can}}\) be the number of retrieved candidate predicates; total space usage is \(O(p_{\text{can}})\).\shorten

\subsection{Query Planning Agent}
Once the Matching Agent produces the mappings of \( \texttt{EntToVertex} \) and \( RelToPred \), the Query Planning Agent uses them along with the \( QIR \) and the user question \( q_i \). It manages SPARQL query construction, selection, and execution through three tools: \textit{Query Generator}, \textit{Query Validator}, and \textit{Query Execution}. Query selection is performed directly by the agent.
As shown in Lines \ref{algo3:line:Q-generate}–\ref{algo3:line:Q-truncate} of Algorithm~\ref{alg:query_selection}, the Query Generator creates a candidate set \( \mathcal{Q} = \{ \psi_1, \dots, \psi_k \} \) by combining entity mappings from \( \texttt{EntToVertex} \) with predicate options from \( \texttt{RelToPred} \). To limit latency, the candidate set is truncated to a bounded size \( query\_num \). The agent then filters \( \mathcal{Q} \) based on predicate relevance. Before execution, selected queries are passed to the \textit{Query Validator}, which checks structural correctness and semantic validity. Finally, the validated queries \( \mathcal{Q'} \) are executed via the SPARQL endpoint using the Query Execution tool, and the retrieved answers are aggregated into the final result \( A \), which is returned to the Chat Agent.

\subsubsection{Query Generator}
The Query Generator constructs a candidate set of SPARQL queries \( \mathcal{Q} = \{ \psi_1, \dots, \psi_k \} \) using the QIR as the structural backbone. It populates each triple \( \langle s, r, o \rangle \in RF \) by replacing entities with their corresponding vertices from \texttt{EntToVertex}, and substituting relations with predicate options from \texttt{RelToPred}($r$). Unknowns are included directly as variables.
Since each relation may map to multiple predicates, a single semantic triple can yield multiple SPARQL triple patterns. The final queries are formed by joining these patterns based on shared variables or entities. This combinatorial expansion produces multiple candidate queries, each representing a distinct predicate configuration. The main unknown representing the target answer is added to the \texttt{SELECT} clause.

\subsubsection{Query Selection}
Lines \ref{algo3:line:P_PredToQuery_init}–\ref{algo3:line:Q_dash} of Algorithm~\ref{alg:query_selection} describe the query selection process performed by the Query Planning Agent. The agent iterates over the candidate queries \( \psi \in \mathcal{Q} \), extracts their predicate sets \( \texttt{Pred}(\psi) \), and aggregates them into a global list \( \mathcal{P} \). Simultaneously, it builds an inverted index \( \texttt{PredToQuery} \) that maps each predicate to the queries containing it. To reduce execution cost and filter out irrelevant queries, the agent applies an LLM-based predicate filtering strategy. It provides the user question \( q_i \), predicate list \( \mathcal{P} \), and a zero-shot chain-of-thought prompt \( I_{QS} \) to the LLM. The model returns a filtered subset \( \mathcal{P'} \subseteq \mathcal{P} \) of predicates considered relevant to the question. This output is validated using the Query Validator to ensure it is well-formed JSON and contains only predicates from the original list. The process is retried up to \( \theta \) times if the output is invalid or malformed. The agent then filters the candidate query set using the validated predicate subset: \( \mathcal{Q'} = \{ \psi \in \mathcal{Q} \mid \texttt{Pred}(\psi) \cap \mathcal{P'} \neq \emptyset \} \). This step reduces query volume while preserving recall, which help improve efficiency without sacrificing answer accuracy.

\begin{algorithm}[t]
\caption{Query Selection and Execution}
\label{alg:query_selection}
\begin{flushleft}
\textbf{Input:} $QIR$: Intermediate representation, $EntToVertex$: Entity-to-vertex map, $RelToPred$: Relation-to-predicate map, $query\_num$: Query limit, $q_i$: User question, $endpoint$: SPARQL endpoint, $I_{QS}$: Predicate selection prompt, $\theta$: Retry threshold\\
\textbf{Output:} $A$: Final answer
\end{flushleft}
\begin{algorithmic}[1]
\STATE $\mathcal{Q} \gets \text{QPAgent.Gen}(QIR, EntToVertex, RelToPred)$
\label{algo3:line:Q-generate}
\STATE $\mathcal{Q} \gets \text{Truncate}(\mathcal{Q}, query\_num)$
\label{algo3:line:Q-truncate}
\STATE $\mathcal{P}, \texttt{PredToQuery} \gets [], \{\}$
\label{algo3:line:P_PredToQuery_init}
\FOR{each $i, \psi \in \mathcal{Q}$}
    \STATE $P_{\psi} \gets \text{QPAgent.Pred}(\psi)$
    \STATE $\mathcal{P} \gets \mathcal{P} \cup P_{\psi}$
    \FOR{each $p \in P_{\psi}$}
        \STATE $\texttt{PredToQuery}[p] \gets \texttt{PredToQuery}[p] \cup \{i\}$
    \ENDFOR
\ENDFOR
\label{algo3:line:for_end}
\STATE $\text{valid} \gets \text{False}, \theta \gets \theta$
\label{algo3:line:valid}
\WHILE{not valid and $\theta > 0$}
    \STATE $\mathcal{P'} \gets \text{QPAgent}(q_i, \mathcal{P}, I_{QS})$ \hfill
    \STATE $\text{valid} \gets \text{QPAgent.Validate}(\mathcal{P'}, \mathcal{P})$ \hfill
    \STATE $\theta \gets \theta - 1$
\ENDWHILE
\label{algo3:line:while_end}
\STATE $\mathcal{Q'} \gets \{ \psi \in \mathcal{Q} \mid \texttt{QPAgent.Pred}(\psi) \cap \mathcal{P'} \neq \emptyset \}$
\label{algo3:line:Q_dash}
\STATE $A \gets \text{QPAgent.Exec}(\mathcal{Q'}, endpoint)$
\STATE \textbf{return} $A$
\end{algorithmic}
\end{algorithm}

\vspace*{-1ex}
\subsubsection{Query Validator}
The Query Validator tool verifies the selected predicates \( \mathcal{P'} \) by ensuring that (i) the LLM output is a well-formed, parsable JSON object, and (ii) \( \mathcal{P'} \) contains only predicates from the candidate set \( \mathcal{P} \). The validator filters out any predicate in \( \mathcal{P'} \) not found in \( \mathcal{P} \). If the resulting list is empty, the output is deemed invalid, and the Query Planning Agent retries the selection.
Entities are already grounded by the Matching Agent, and SPARQL queries are constructed deterministically from the selected predicates. Therefore, validating \( \mathcal{P'} \) effectively guarantees the correctness of the final queries. These checks mitigate LLM hallucinations with minimal overhead. 
Since \( \mathcal{P} \) is stored locally as a hash-based structure, validation is efficient. 

\vspace*{-1ex}
\subsubsection{Query Execution}
The Query Execution tool dispatches the filtered SPARQL queries \( \mathcal{Q'} \) to the knowledge graph and aggregates the results into a unified answer set \( A \). It interfaces with a standard SPARQL endpoint using API calls, executes each query in \( \mathcal{Q'} \), collects result bindings, removes duplicates, and formats the final output for the Chat Agent. This component is lightweight, stateless, and compatible with standard RDF engines.

\subsubsection{Error Handling and Fallback Strategy}
If no answer is returned, this indicates that the KG may not contain the requested information. The system does not attempt to generate an alternative answer, as this may lead to hallucination. This design ensures the system returns either a factually grounded response based on available data or explicitly indicates that the requested information is not present in the KG. Inconsistent results are avoided through assertion-based validations applied throughout the pipeline, especially during query selection. These validations ensure that intermediate outputs remain relevant to the user’s question.

\noindent \textbf{Time and Space Complexity}
Algorithm~\ref{alg:query_selection} has a time complexity of \( O(q_{\text{cost}} + \theta \cdot C_{\text{LLM}}) \), where \( q_{\text{cost}} \) denotes the cost of SPARQL query execution and \( C_{\text{LLM}} \) is the cost of a single LLM call. The loop over candidate queries (Lines \ref{algo3:line:P_PredToQuery_init}–\ref{algo3:line:for_end}) is bounded by the truncation parameter \( query\_num \), typically less than 100, and does not affect asymptotic complexity. LLM output validation (Lines \ref{algo3:line:valid}–\ref{algo3:line:while_end}) incurs constant time per attempt, with up to \( \theta \) retries. 
Space complexity is dominated by the storage of candidate queries and the predicate-to-query index. Both scale linearly with \( query\_num \), resulting in overall space complexity of \( O(query\_num) \). 
Since fewer queries are usually executed than \(query\_num\), the algorithm keeps overhead low while preserving accuracy and robustness.

\section{Implementation}
We implement \sysName{}
\footnote{\url{https://github.com/CoDS-GCS/Chatty-KG}}
using the LangGraph framework\footnote{\url{https://python.langchain.com/docs/langgraph/}}, an extension of LangChain that introduces graph-based abstractions for multi-agent workflows. LangGraph enables structured control over agent composition and inter-agent communication via a shared state and programmable routing logic.
%
Each agent in {\sysName} is implemented as an independent function that reads and updates a shared \texttt{AgentState} object. This object acts as centralized memory, storing the user question, intermediate outputs, and final results. These agent functions are composed into a \texttt{StateGraph}, where conditional transitions depend on the evolving state.
As shown in Listing~\ref{lst:langgraph-assembly}, execution begins at \texttt{chat\_agent} (Line~19) and proceeds through agent nodes (Lines~12–17), including \texttt{classifier\_agent}, \texttt{rephraser\_agent}, and \texttt{qir\_agent}. While Figure~\ref{fig:framework-agent} depicts a fixed logical view, the implementation is dynamic: each agent sets the \texttt{route} field to determine the next node. This decouples routing from static topology, enabling context-sensitive transitions and modular reconfiguration without altering control flow.

\begin{figure}[t]
 \vspace*{-4ex}
\centering
\begin{lstlisting}[%style=chattykg, 
language=Python, label={lst:langgraph-assembly}, caption={LangGraph-based implementation of {\sysName}. The shared agent state tracks intermediate outputs, such as rephrased questions, SPARQL queries, and routing decisions, exchanged among agents during dialogue processing.}]
# Agent function signatures (each takes and returns AgentState)
def chat_agent(state: AgentState) -> AgentState: ...
def classifier_agent(state: AgentState) -> AgentState: ...
def rephraser_agent(state: AgentState) -> AgentState: ...
def qir_agent(state: AgentState) -> AgentState: ...
def query_agent(state: AgentState) -> AgentState: ...
def matching_agent(state: AgentState) -> AgentState: ...

# Graph construction
builder = StateGraph(state_schema=AgentState)

builder.add_node("chat_agent", chat_agent)
builder.add_node("classifier_agent", classifier_agent)
builder.add_node("rephraser_agent", rephraser_agent)
builder.add_node("qir_agent", qir_agent)
builder.add_node("query_agent", query_agent)
builder.add_node("matching_agent", matching_agent)

builder.set_entry_point("chat_agent")

# Define transitions between agent nodes
builder.add_edge("chat_agent", 
   lambda state: END if state.query_done else ("query_agent" if state.qir_done else "qir_agent"))
builder.add_conditional_edges("classifier_agent", 
   lambda state: state.route)
builder.add_conditional_edges("rephraser_agent",  
   lambda state: state.route)
builder.add_conditional_edges("qir_agent",  
   lambda state: state.route)
builder.add_conditional_edges("query_agent",  
   lambda state: state.route)
builder.add_conditional_edges("matching_agent",  
   lambda state: state.route)

builder.set_finish_point("chat_agent")
graph = builder.compile()
\end{lstlisting}
\end{figure}

The modular design of {\sysName} enables adaptability at multiple levels. \textbf{LLM substitution} is straightforward: agents can use different underlying models (e.g., GPT-4, Phi, Mistral) by updating internal configurations. \textbf{Agent extensibility} is also supported, allowing components to be swapped or extended, e.g., replacing a rule-based tool with a neural module, without retraining or reengineering the system. Finally, \textbf{dialogue mode control} is governed by a \texttt{system\_mode} flag within the shared state, allowing the system to toggle between multi-turn and single-turn execution. These features make \sysName{} a flexible and extensible platform for research and deployment in diverse KGQA settings.

{\sysName} uses four main parameters: (1) $\theta$, the maximum number of retries for LLM calls when validation fails; (2) $L$, the number of past answers included in the context $C_q$ to fit within the LLM's input length; (3) $v_{limit}$, the number of candidate vertices retrieved by the entity retriever; and (4) $query\_num$, the maximum number of candidate queries considered during query selection. In our experiments, we set these to $3$, $100$, $600$, and $40$, respectively. The $\theta$ parameter helps reduce hallucinations by allowing the LLM multiple attempts to produce valid outputs. While higher values improve robustness, they also increase cost and latency. Models like GPT-4o and Gemini benefit from retries by producing varied and often valid responses, whereas weaker models like Llama show limited gains.
%
%
We set $L{=}100$ based on empirical testing. Larger values may exceed the LLM context window for long answers (e.g., lists of papers), while smaller values risk omitting context needed by the rephraser.
%
%
The $v_{limit}$ parameter balances recall and prompt complexity: higher values improve recall but add noise, while lower values reduce noise but risk missing the correct vertex. The $query\_num$ value must retain valid queries without increasing runtime. For fairness in evaluation, we disable natural-language reformulation and compare against structured gold-standard answers.

\section{Evaluation}
We evaluate {\sysName} along five key dimensions: (1) performance on single-turn QA compared to state-of-the-art KGQA systems, (2) accuracy on multi-turn questions over KGs relative to top-performing chatbots, LLMs and RAG-based systems, (3) the contribution of our contextual understanding module, (4) the effect of our LLM-powered question interpretation, and (5) the impact of our query planning and selection on system efficiency.


\begin{table}[t]
\vspace*{-2ex}
  \caption{KG Statistics: The number of triples and entities in Millions, and the number of predicates of each KG.
  }
  \vspace*{-3ex}
  \label{tab:stats}
  \begin{tabular}{cccc}
    \toprule
    \textbf{KG} & \textbf{\# Entities(M)} & \textbf{\# Triples(M)}&\textbf{\# Predicates}\\
    \midrule
    \textbf{DBLP}&18& 263&167\\
    \textbf{YAGO} & 12&207& 259\\
     \textbf{DBPedia} & 14 & 350&60,736\\
     \textbf{MAG} & 586 &13, 705&178\\
     \textbf{Wikidata\footnotemark} &119   &8,541 & 12,943 \\

   \bottomrule
\end{tabular}
\vspace*{-4ex}
\end{table}
\footnotetext{\url{https://www.wikidata.org/wiki/Wikidata:Statistics}}


\subsection{Evaluation Setup}

\textbf{Baselines:} 
We evaluate {\sysName} against state-of-the-art systems in two settings: single-turn and multi-turn question answering. For single-turn, we compare against KGQAn~\cite{kgqan}, a leading KGQA system, and EDGQA~\cite{EDGQA}. For multi-turn (conversational) QA, we compare against CONVINSE~\cite{convinse} and EXPLAIGNN~\cite{expalignn}. Both are designed for KG-based dialogue and trained on the ConvMix dataset~\cite{convinse}, which augments multi-turn QA over Wikidata. Unlike {\sysName}, these systems require task-specific training and are tightly coupled to the KG they were trained on. 
%
We evaluated two widely used \textbf{\textit{RAG systems}}, ColBERT~\cite{santhanam2022colbertv2, khattab2020colbert} and GraphRAG~\cite{graph_rag}, on DBLP (our smallest KG). Both require substantial compute and memory. ColBERT performs late-interaction retrieval without building a graph, while GraphRAG constructs and indexes one for downstream QA. Indexing DBLP with GraphRAG required over 80 hours using a 512\,GB, 64-core VM, used solely for the RAG experiment. Extrapolating from this, larger KGs would require terabytes of memory, making RAG indexing impractical in our setting.

\noindent \textbf{Underlying LLMs:} We evaluate {\sysName} using: \myNum{1} Commercial LLMs, GPT-4o~\cite{gpt4o}, Gemini-2.0-Flash~\cite{geminiteam2024gemini15}, and DeepSeek-V3~\cite{deepseekv3}, accessed via API. \myNum{2} Open-weight LLMs, such as  Qwen-2.5~\cite{qwen2.5}, Phi-4~\cite{phi4}, Gemma-3~\cite{gemma3}, CodeLlama~\cite{codellama}, Gemma-2~\cite{gemma2}, Granite~\cite{granite}, Vicuna~\cite{vicuna}, Llama-3~\cite{llama3.1}, and Mistral~\cite{mistral}, all hosted locally.\shorten

\begin{table*}[t]
\vspace*{-2ex}
    \fontsize{7pt}{7pt}\selectfont 
  \caption{
  Comparison between state-of-the-art KGQA systems and {\sysName} on single-turn question answering across five benchmarks covering four knowledge graphs. {\sysName} is evaluated with various LLMs. Open-weight LLMs are ordered top-down by parameter count. Bold indicates best performance, underlined indicates second-best. Green highlights performance better than KGQA systems; yellow highlights results within 1 point. RAG systems require huge memory to run.
  }
  \vspace*{-1ex}
  \label{tab:differentllms}
  \begin{tabular}{lccc|ccc|ccc|ccc|ccc}
    \toprule
      & \multicolumn{3}{c|}{\textbf{{\qald}}} & \multicolumn{3}{c|}{\textbf{{\lcq}}} & 
       \multicolumn{3}{c|}{\textbf{{YAGO}}} & \multicolumn{3}{c|}{\textbf{{DBLP}}} & \multicolumn{3}{c}{\textbf{{MAG}}}  \\
     \textbf{System(Model)}&{\textbf{P}} &{\textbf{R}} &{\textbf{F1}} &{\textbf{P}} &{\textbf{R}} &{\textbf{F1}} 
        &{\textbf{P}} &{\textbf{R}} &{\textbf{F1}} &{\textbf{P}} &{\textbf{R}} &{\textbf{F1}} &{\textbf{P}} &{\textbf{R}} &{\textbf{F1}} \\
    \midrule
        \textbf{{\kgqan}}    & 51.13  & 38.72 &  44.07   & 58.71   &  \underline{46.11}   &  51.65 &   
        48.48  & 65.22&  55.62 &  57.87    &   52.02   & 54.79 &   55.43   & 45.61  &    50.05\\
        \textbf{EDGQA}&31.30&40.30&32.00&50.50&\textbf{56.00}&53.10&41.90&40.80&41.40&8.00&8.00&8.00&4.00&4.00&4.00\\

        \midrule
        \textbf{GraphRAG  (GPT-4o)}    & -  & - &  -   & -   &  -   &  - &   
        -    &   -   & - & 46.00  & 37.01 &  38.99 &     -   & -  &   -\\

        \textbf{COLBERT (GPT-4o)}    & -  & - &  -   & -   &  -   &  - &   
        -    &   -   & - & 65.38 & 55.33 &  55.29 &     -   & -  &   -\\

        \midrule
        \textbf{\sysName}(\textbf{GPT-4o})&\textbf{61.91}&\underline{41.50}&\cellcolor{green!30}\textbf{49.69}&\textbf{72.79}&43.60&\cellcolor{green!30}\underline{54.54}&86.52&72.27&\cellcolor{green!30}78.75&\underline{85.96}&75.50&\cellcolor{green!30}\textbf{80.39}&\textbf{75.05}&\underline{69.05}&\cellcolor{green!30}\textbf{71.92}\\
        \textbf{\sysName}(\textbf{Gemini-2.0-Flash})&\underline{57.23}&\textbf{42.14}&\cellcolor{green!30}\underline{48.54}&68.27&43.35&\cellcolor{yellow}53.03&\textbf{90.04}&\textbf{75.47}&\cellcolor{green!30}\textbf{82.11}&74.80&67.50&\cellcolor{green!30}70.96&59.55&57.56&\cellcolor{green!30}58.54\\ 
        \textbf{\sysName}(\textbf{DeepSeek-V3})&50.76&40.83&\cellcolor{green!30}45.26&\underline{69.66}&45.42&\cellcolor{green!30}\textbf{54.99}&78.77&\underline{74.07}&\cellcolor{green!30}76.34&71.88&\textbf{86.50}&\cellcolor{green!30}\underline{78.52}&64.43&\textbf{72.36}&\cellcolor{green!30}\underline{68.16}\\
        \midrule
        \textbf{\sysName}(\textbf{Phi-4})&55.95&37.75&\cellcolor{green!30}45.08&69.09&42.43&\cellcolor{yellow}52.57&\underline{86.89}&\underline{74.07}&\cellcolor{green!30}\underline{79.97}&81.04&75.50&\cellcolor{green!30}78.17&63.23&62.70&\cellcolor{green!30}62.96\\
        \textbf{\sysName}(\textbf{Qwen2.5-14B-Inst})&50.77&42.17&\cellcolor{green!30}46.07&68.84&42.69&\cellcolor{yellow}52.70&83.73&67.07&\cellcolor{green!30}74.48&65.96&73.50&\cellcolor{green!30}69.53&57.58&61.93&\cellcolor{green!30}59.68\\
        \textbf{\sysName}(\textbf{Qwen2.5-14B})&48.99&39.84&\cellcolor{yellow}43.94&62.49&41.29&49.72&74.77&71.76&\cellcolor{green!30}73.23&58.61&68.50&\cellcolor{green!30}63.17&61.73&62.56&\cellcolor{green!30}62.14\\
        \textbf{\sysName}(\textbf{Vicuna-13B-v1.5})&62.40&26.17&36.87&73.78&27.92&40.51&58.53&36.00&44.58&64.90&30.50&41.50&75.93&34.51&47.45\\
        \textbf{\sysName}(\textbf{CodeLlama-13B})&43.99&32.32&37.26&56.31&38.41&45.67&44.48&53.24&48.47&39.08&68.00&49.64&34.86&61.82&44.58\\
        \textbf{\sysName}(\textbf{Gemma3-Inst})&53.64&36.93&\cellcolor{yellow}43.75&66.13&39.09&49.13&72.67&65.87&\cellcolor{green!30}69.10&71.44&\underline{76.50}&\cellcolor{green!30}73.88&49.28&43.52&46.22\\
        \textbf{\sysName}(\textbf{Gemma3})&57.11&34.76&\cellcolor{yellow}43.22&63.65&39.33&48.62&75.33&72.79&\cellcolor{green!30}74.04&57.06&63.50&\cellcolor{green!30}60.11&54.64&51.30&\cellcolor{green!30}52.92\\
        \textbf{\sysName}(\textbf{Gemma2-Inst})&51.16&35.47&41.90&65.88&39.85&49.66&81.57&60.47&\cellcolor{green!30}69.45&79.81&68.00&\cellcolor{green!30}73.43&\underline{65.95}&49.22&\cellcolor{green!30}56.37\\
        \textbf{\sysName}(\textbf{Gemma2})&43.19&37.59&40.20&53.95&41.56&46.95&62.93&66.47&\cellcolor{green!30}64.65&48.46&58.00&52.80&48.19&44.23&46.12\\
        \textbf{\sysName}(\textbf{Ministral-8B-Inst})&56.72&37.47&\cellcolor{green!30}45.13&65.31&39.98&49.60&78.92&72.19&\cellcolor{green!30}75.41&\textbf{87.94}&69.50&\cellcolor{green!30}77.64&58.17&58.54&\cellcolor{green!30}58.35\\
        \textbf{\sysName}(\textbf{Granite-3.2-8B-Inst})&47.23&36.68&41.29&66.59&34.27&45.25&69.68&72.46&\cellcolor{green!30}71.04&74.09&63.00&\cellcolor{green!30}68.10&51.00&56.71&\cellcolor{green!30}53.70\\
        \textbf{\sysName}(\textbf{Llama-3.1-8B-Inst})&56.31&32.54&41.25&65.66&37.63&47.84&67.61&70.66&\cellcolor{green!30}69.10&66.24&60.01&\cellcolor{green!30}62.98&63.87&64.06&\cellcolor{green!30}63.97\\
        \textbf{\sysName}(\textbf{Mistral-7B-v0.3})&48.75&36.02&41.43&61.39&33.41&43.27&50.02&65.99&\cellcolor{green!30}56.91&47.36&63.00&\cellcolor{yellow}54.07&35.09&59.28&44.09\\
        \bottomrule
  \end{tabular}
\end{table*}

\noindent \textbf{Five Real-World KGs:}
To assess {\sysName}'s performance across domains, we evaluate it on 
five real-world KGs: DBpedia~\cite{dbpedia}, YAGO~\cite{YAGOkgdownload, yago}, DBLP~\cite{DBLPrelease}, Microsoft Academic Graph (MAG)~\cite{MAGrecords}, and Wikidata~\cite{wikidata}. DBpedia, Wikidata and YAGO cover general knowledge (e.g., people, places), while DBLP and MAG focus on academic content, including long entity names like paper titles. These KGs vary in size, enabling evaluation of {\sysName}'s scalability and robustness. Table~\ref{tab:stats} summarizes the number of entities, predicates, and triples. For wikidata SPARQL execution, we use the public available endpoint. For the other graph's SPARQL execution, we use Virtuoso v7.2.5.2, a standard backend for large KGs, with a separate endpoint per KG. All experiments use the default Virtuoso setup.\shorten

\noindent \textbf{Compute Infrastructure:}
We use two different setups for our experiments: \myNum{i} Linux machine with 16 cores and 180GB RAM for running all experiments on {\sysName} and {\kgqan}. It is also used to host the four Virtuoso SPARQL endpoints for the KGs. \myNum{ii} Linux machine with Nvidia A100 40GB GPU, 32 cores, and 64GB RAM to host the open-weight LLMs, creating a server accessible using HTTP requests via the VLLM library \cite{vllm}.

\noindent \textbf{Benchmarks:}
We evaluate {\sysName} on a comprehensive set of KGQA benchmarks spanning both single-turn and multi-turn QA over multiple KGs. For \textbf{single-turn QA}, we use standard datasets such as {\qald}~\cite{qald9} and {\lcq}~\cite{lcquad}, which provide SPARQL-annotated questions and gold answers over DBpedia. QALD-9 includes 408 training and 150 test questions, while LC-QuAD 1.0 provides 4,000 training and 1,000 test questions generated from diverse templates. Since {\sysName} requires no training, we only use the test sets. We also include datasets introduced by~\cite{kgqan}, with 100 questions each for YAGO, DBLP, and MAG, covering varied domains and KG structures.
For \textbf{multi-turn QA}, we generate 20 dialogues per KG (approximately 100 questions) using a state-of-the-art dialogue benchmark generator~\cite{chattygen}. Each dialogue consists of 5 turns, with every question $q$ paired with its standalone version, a corresponding SPARQL query, and the correct answer. These generated benchmarks allow us to evaluate {\sysName}'s contextual understanding and reasoning capabilities in a reproducible setting.\shorten

\noindent \textbf{Metrics:} For single-turn QA, we use standard KGQA metrics: Precision (P), Recall (R), and F1 score. Recall measures the proportion of correct answers returned by the system, while Precision measures the proportion of returned answers that are correct. The F1 score is the harmonic mean of Precision and Recall.
For multi-turn QA, we adopt the metrics used by CONVINSE and EXPLAIGNN: Precision at 1 (P@1), Mean Reciprocal Rank (MRR), and Hit at 5 (Hit@5). P@1 checks whether the top-ranked answer is correct and is common when systems return a single answer~\cite{convinse}. MRR evaluates the position of the correct answer in a ranked list. Hit@5 measures whether the correct answer appears among the top five results.

\begin{table*}[t]
\vspace*{-2ex}
 \caption{
 Performance of {\sysName} on multi-turn questions, compared to conversational baselines (CONVINSE, EXPLAIGNN) and general LLMs (GPT-4o, Gemini, DeepSeek, Phi-4, Qwen) across three KGs. Metrics: P@1, MRR, and Hit@5. {\sysName} outperforms all baselines, achieving up to +87\% higher P@1 on YAGO and nearly 3$\times$ higher accuracy on DBLP. 
 }
 \vspace*{-2ex}
 \label{tab:baseline}
 \centering
 \begin{tabular}{l|ccc|ccc|ccc|ccc}
    \toprule
    \textbf{System} & \multicolumn{3}{c|}{\textbf{Wikidata}} & \multicolumn{3}{c|}{\textbf{DBpedia}} & \multicolumn{3}{c|}{\textbf{YAGO}} & \multicolumn{3}{c}{\textbf{DBLP}} \\
    & P@1 & MRR & Hit@5 & P@1 & MRR & Hit@5 & P@1 & MRR & Hit@5 & P@1 & MRR & Hit@5 \\
    \midrule
    \textbf{CONVINSE} & 27.47&27.47 & 27.47&28.72 &28.72 & 28.72& 34.88& 34.88& 34.88& 28.89&28.89 &28.89 \\
    \textbf{EXPLAIGNN} & 26.37 & 26.37&26.37 &29.79 &29.79 & 29.79& 31.40&31.40 &31.40 &28.89 &28.89 &28.89 \\
    \midrule
    \textbf{GraphRAG (GPT-4o)}& - & - & - & - & - & - & - & - & - & 11.10 & 11.10 & 11.10 \\
    \textbf{ColBERT (GPT-4o)} & - & - & - & - & - & - & - & - & - & 17.78 & 17.78 & 17.78 \\
    \midrule
    \textbf{GPT-4o} & 49.45 & 49.45& 49.45 & 30.85& 30.85& 30.85& 27.91&27.91 &30.23 &24.44 &24.44 &24.44 \\
    \textbf{Gemini-2.0-Flash} & 41.76& 41.76 & 41.76& 25.53& 24.47 & 24.47 & 30.23 &31.10 &32.56 & 13.33&13.33&13.33\\
    \textbf{DeepSeek-V3} & 40.65 &40.65 &40.65 &20.21 &20.21 &20.21 &26.74 &26.74 &26.74 &16.67 &16.67 &16.67 \\
    \midrule
    \textbf{Phi-4} & 25.27 & 25.27 & 25.27 & 26.60& 25.53& 25.53& 15.12&15.31 &15.12 & 17.78&17.78 &17.78 \\
    \textbf{Qwen2.5-14B-Instruct} & 24.18 & 24.18 & 24.18 & 18.09& 18.09& 18.09& 16.28& 16.86&17.44 & 6.67&5.56 &5.56 \\
    \midrule
    \textbf{{\sysName}(GPT-4o)} & \textbf{54.95} & \textbf{54.95} &\textbf{54.95} & \textbf{34.04}& \textbf{35.11}& \textbf{36.17}& \textbf{65.12}&\textbf{65.70} &\textbf{66.28} &\textbf{83.33} &\textbf{83.33} &\textbf{83.33} \\
    \bottomrule
  \end{tabular}
  \vspace*{-2ex}
\end{table*}

\subsection{Single-Turn KGQA}
We evaluate {\sysName} on single-turn QA using five benchmarks and compare it with two state-of-the-art systems: KGQAn and EDGQA. Results are reported in Table~\ref{tab:differentllms}. We also assess how {\sysName}'s performance varies when paired with different commercial and open-weight LLMs. {\sysName} outperforms both baselines when using high-performing LLMs like GPT-4o, Gemini-2.0, and DeepSeek. For example, GPT-4o achieves F1 scores of 80.39 on DBLP and 71.92 on MAG, compared to KGQAn's 54.79 and 50.05. On QALD and LC-QuAD, it leads with F1 scores of 49.69 and 54.54, improving over KGQAn by +5.62 and +2.89, respectively.

GPT-4o also yields strong precision, such as 86.52 on YAGO, showing that {\sysName}'s query planning effectively filters incorrect queries. DeepSeek-V3 achieves high recall, e.g., 86.50 on DBLP. Models like Phi-4 and Qwen2.5-14B-Instruct provide a good trade-off between precision and recall, outperforming KGQAn in the five benchmarks. 
Notably, even smaller open-weight models like Mistral-7B and Llama-3.1-8B match or exceed KGQAn on some datasets. This shows that {\sysName} achieves strong performance without task-specific fine-tuning. Our modular prompting and accurate query selection enable this capability, even with smaller models.\shorten

EDGQA performs competitively on LC-QuAD (F1 = 53.10) but fails on domain-specific KGs, such as DBLP (F1 = 8.0) and MAG (F1 = 4.0), likely due to its rule-based architecture. KGQAn performs better overall but struggles with precision on general datasets like {\qald} (P = 51.13) and YAGO (P = 48.48), often retrieving irrelevant answers. In contrast, {\sysName} maintains high precision across both general and special domains. It delivers strong F1 and precision scores on datasets like DBLP and YAGO. These results demonstrate the effectiveness of our query planning agent. Given GPT-4o's consistent performance across all datasets and metrics, we use GPT-4o as the default model for the remainder of our evaluation, unless otherwise specified.\shorten


We also evaluated two RAG systems, GraphRAG and ColBERT, to contextualize our results. On DBLP, GraphRAG achieved moderate precision (46.00) but low recall (37.01), while ColBERT performed better (P = 65.38, R = 55.33, F1 = 55.29). Both struggled with list-style questions, often returning only partial answers, and failed on multi-hop queries because text chunking breaks graph structure. GraphRAG also required heavy indexing and still missed relations, confirming the scalability challenge. These results support our observation that current RAG implementations cannot reliably preserve structure or guarantee complete answers. In contrast, {\sysName} operates directly over the KG and executes SPARQL, enabling accurate, complete retrieval and consistent performance.

\subsection{Multi-turn Dialogues}
We evaluate {\sysName} on multi-turn question answering using 20 dialogues (up to 100 questions) generated by a state-of-the-art benchmark generator~\cite{chattygen}. We compare against two conversational KGQA systems, CONVINSE and EXPLAIGNN, and several general-purpose LLMs 
that do not have access to KGs
(GPT-4o, Gemini-2.0-Flash, DeepSeek-V3, Phi-4, and Qwen2.5-14B-Instruct).

\noindent \textbf{Dialogue Datasets.}
CONVINSE and EXPLAIGNN were trained on Wikidata, while we evaluate on DBpedia, YAGO, and DBLP. To ensure fairness, we manually selected 20 entities that exist in both Wikidata and our target KGs. For DBpedia and YAGO, we verified alignment using Wikidata links. For DBLP, which lacks clear mapping, we selected entities relevant to research discussions (e.g., “Gerhard Kramer”). CONVINSE and EXPLAIGNN were accessed through their official demos\footnote{https://convinse.mpi-inf.mpg.de/}\footnote{https://explaignn.mpi-inf.mpg.de/}, and their outputs were converted to match our evaluation format. The performance of CONVINSE and EXPLAIGNN in our setup matches the results reported in their original papers, confirming the fairness of our experimental design.

\noindent \textbf{Performance Results.}
Table~\ref{tab:baseline} presents results using P@1, MRR, and Hit@5. {\sysName} outperforms all baselines and LLMs across every KG and metric. 
On Wikidata, it achieves 54.95 P@1, significantly outperforming CONVINSE (27.47) and EXPLAIGNN (26.37) which were trained on Wikidata. 
On YAGO, it achieves an 87\% performance gain in P@1 compared to CONVINSE.
On DBLP, a domain-specific KG, it reaches 83.33 P@1, nearly 3X the score of the best baseline. On DBpedia, {\sysName} still leads with 34.04 P@1, ahead of EXPLAIGNN's 29.79. 
%
%
The improvement on DBpedia is smaller due to its scale and structure. As shown in \autoref{tab:stats}, DBpedia has 60{,}736 predicates, about $234\times$ more than YAGO and $4\times$ more than Wikidata, which introduces high ambiguity in predicate selection. Many predicates are also redundant or synonymous (e.g., \textit{founders}, \textit{founder}, \textit{foundedBy} for \textit{Intel}), making query planning more difficult. Despite this, {\sysName} remains competitive on DBpedia and does so without KG-specific training or preprocessing.
%
Our modular agents ground responses in the KG through accurate linking, context reasoning, and minimal query execution. This design adapts to different graph structures and remains robust across domains without retraining or pipeline changes.

\begin{table}[t]
\caption{
Performance comparison of {\sysName} with and without the conversational module using GPT-4o. Dialogue results reflect context-dependent questions handled with the module, while Standalone results use context-free questions. Retention (\% = F1\textsubscript{Dialogue} / F1\textsubscript{Standalone} $\times$ 100) indicates the proportion of accuracy retained in dialogue settings.
}
\vspace*{-2ex}
\label{tab:dialogue}
\centering
\begin{tabular}{lccccc}
\toprule
\textbf{Dataset} & \textbf{Method} & \textbf{P} & \textbf{R} & \textbf{F1} & \textbf{Retention (\%)} \\
\midrule
\multirow{2}{*}{\textbf{Wikidata}} & Dialogue  & 78.65 & 57.14 & 66.19 & \multirow{2}{*}{\textbf{95.28}} \\
                         & Standalone & 79.75 & 61.54 & 69.47 & \\
\midrule
\multirow{2}{*}{\textbf{DBpedia}} & Dialogue   & 68.19 & 42.55 & 52.40 & \multirow{2}{*}{\textbf{92.27}} \\
                         & Standalone & 69.79 & 47.87 & 56.79 & \\
\midrule
\multirow{2}{*}{\textbf{YAGO}}    & Dialogue   & 74.86 & 67.44 & 70.96 & \multirow{2}{*}{\textbf{85.48}} \\
                         & Standalone & 84.69 & 81.40 & 83.01 & \\
\midrule
\multirow{2}{*}{\textbf{DBLP}}    & Dialogue   & 81.85 & 74.44 & 77.97 & \multirow{2}{*}{\textbf{87.09}} \\
                         & Standalone & 89.07 & 90.00 & 89.53 & \\
\bottomrule
\end{tabular}
\vspace*{-2ex}
\end{table}


We also evaluated two RAG systems, GraphRAG and ColBERT, in the same multi-turn setting. GraphRAG performed poorly (e.g., P@1 = 11.10 on DBLP), and ColBERT showed similarly low results (P@1 = 17.78). Both systems handle each query independently and lack dialogue-state tracking, so they cannot resolve references or preserve context across turns. Their retrieval resets at every step, breaking conversational coherence and leading to low accuracy. In contrast, {\sysName} maintains dialogue context and delivers consistent multi-turn performance.

\noindent \textbf{General LLM Limitations.}
While LLMs used without KG access like GPT-4o and Gemini show moderate performance on DBpedia and YAGO, they struggle on DBLP. For instance, GPT-4o drops to 24.44 P@1 on DBLP, far behind {\sysName}. This is expected, as domain-specific KGs like DBLP may be underrepresented in LLM training data. In contrast, {\sysName} grounds its answers directly in the KG and executes structured queries. This design enables higher accuracy, better verification, and lower latency. It handles evolving domains without retraining or KG-specific pipelines. As LLMs are limited by static training data, {\sysName} provides a more reliable and up-to-date solution for multi-turn KGQA. Its combination of contextual understanding and precise query planning ensures high accuracy and low latency without requiring retraining.

\subsection{Analyzing {\sysName}}

\noindent \textbf{Contextual Understanding:}
We evaluate {\sysName}'s ability to handle multi-turn questions. For this, we reuse the same benchmarks from the multi-turn evaluation with CONVINSE and EXPLAIGNN. Each dialogue in these benchmarks includes aligned question pairs: a context-dependent version and its corresponding standalone form. 
We run {\sysName} in two modes: (1) with the Classifier and Rephraser Agents enabled, using the context-dependent questions; and (2) with these agents disabled, using the standalone versions. This setup tests the system’s ability to reconstruct missing context from dialogue history.
As shown in Table~\ref{tab:dialogue}, {\sysName} retains over 85\% of its standalone F1 score across all datasets. This demonstrates the effectiveness of the Classifier and Rephraser Agents in resolving contextual dependencies with minimal performance loss. The slight drop reflects the inherent ambiguity in multi-turn inputs, which {\sysName} mitigates effectively.\shorten

\begin{figure}[t]
\vspace*{-2ex}
  \centering
  \includegraphics[width=\columnwidth]{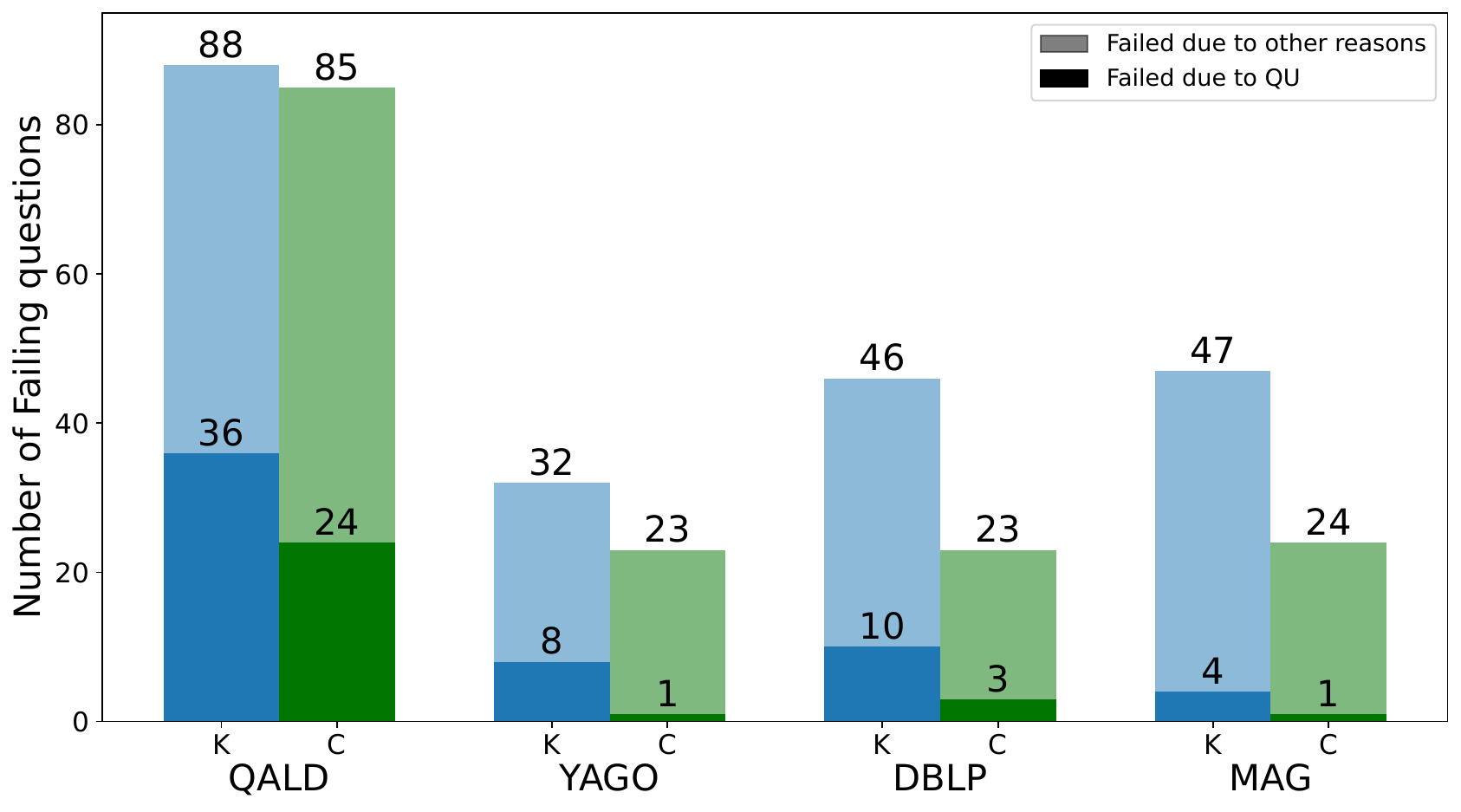}
  \vspace*{-5ex}
  \caption{Number of failed questions (i.e., Recall = 0) in each benchmark. Each bar is divided to show failures caused by Question Understanding (QU) and other factors. Shading variations indicate the source of failure. {\sysName} consistently has the fewest failures across all benchmarks.}
  \label{fig:question_understanding}
  \vspace*{-3ex}
\end{figure}

\noindent \textbf{Question Understanding:}
We evaluate {\sysName}'s ability to understand questions by analyzing failure cases, as shown in Figure~\ref{fig:question_understanding}. The figure compares the number of failing questions (Recall=0) for {\sysName} (denoted as \textbf{C}) and {\kgqan} (denoted as \textbf{K}) across four benchmarks: QALD, YAGO, DBLP, and MAG. Each bar is split into two segments: failures due to \textbf{question understanding (QU)} and failures due to \textbf{other reasons}. Darker shades indicate QU-related failures, while lighter shades capture all other causes.
Across all benchmarks, {\sysName} (C) shows significantly fewer failures than {\kgqan} (K), especially in question understanding. For example, in the YAGO and MAG datasets, {\sysName} fails due to QU in only one case, compared to eight and four cases, respectively, for {\kgqan}. This demonstrates that {\sysName}'s LLM-based modular agents are more robust in interpreting diverse and complex linguistic questions, outperforming {\kgqan}'s fine-tuned model.


\begin{figure}[t]
\vspace*{-2ex}
  \centering
  \includegraphics[width=\columnwidth]{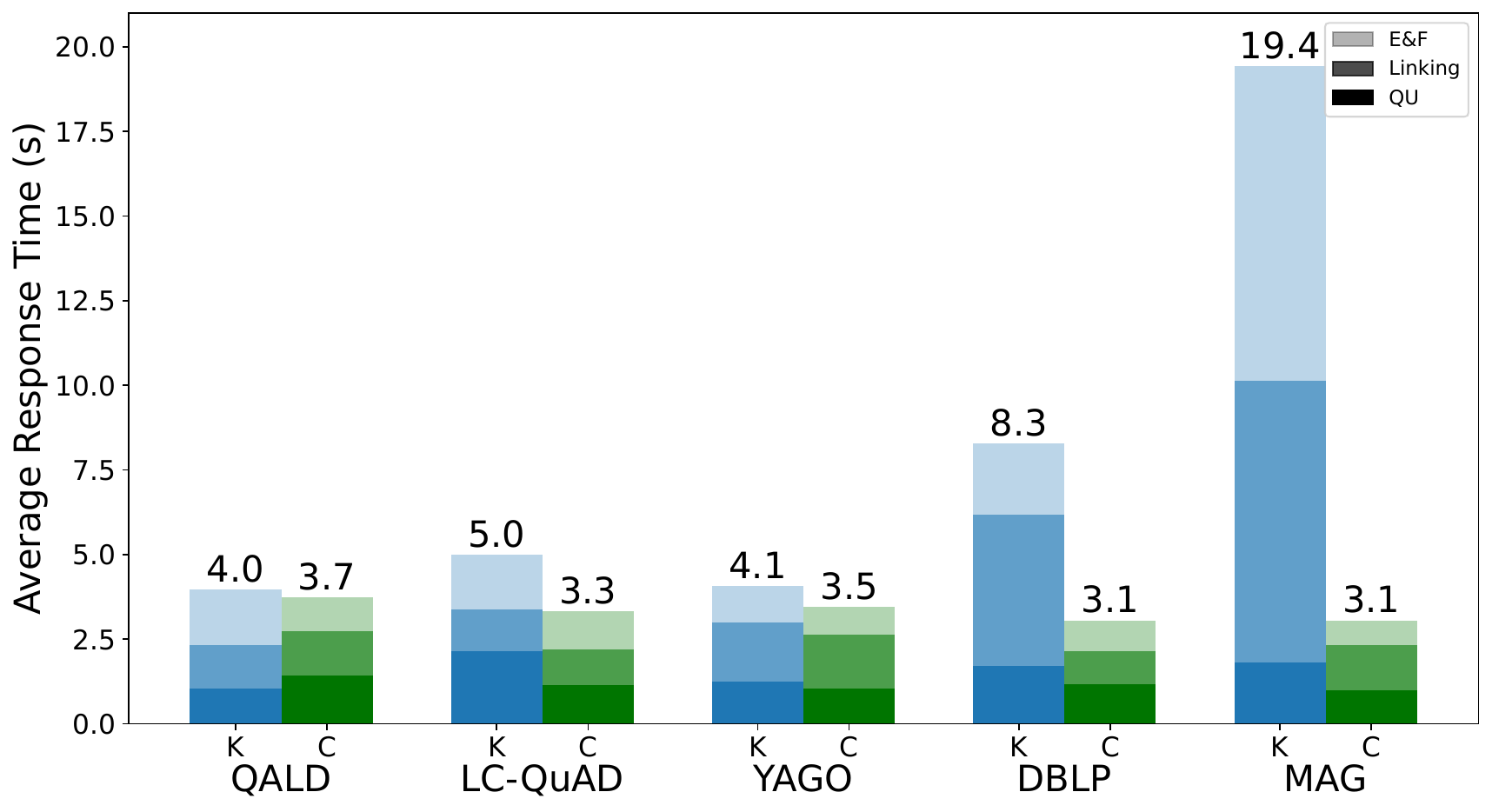}
  \vspace*{-5ex}
  \caption{Average response time per question (in seconds) for {\kgqan} (K) and {\sysName} (C). Each bar is segmented bottom-up into three stages: Question Understanding (QU), Linking, and Execution \& Filtration (E\&F). Shading variations within each bar indicate the contribution of each stage.}

  \label{fig:responsetime}
  \vspace*{-3ex}
\end{figure}

\begin{figure}[t]
  \centering
  \includegraphics[width=\columnwidth]{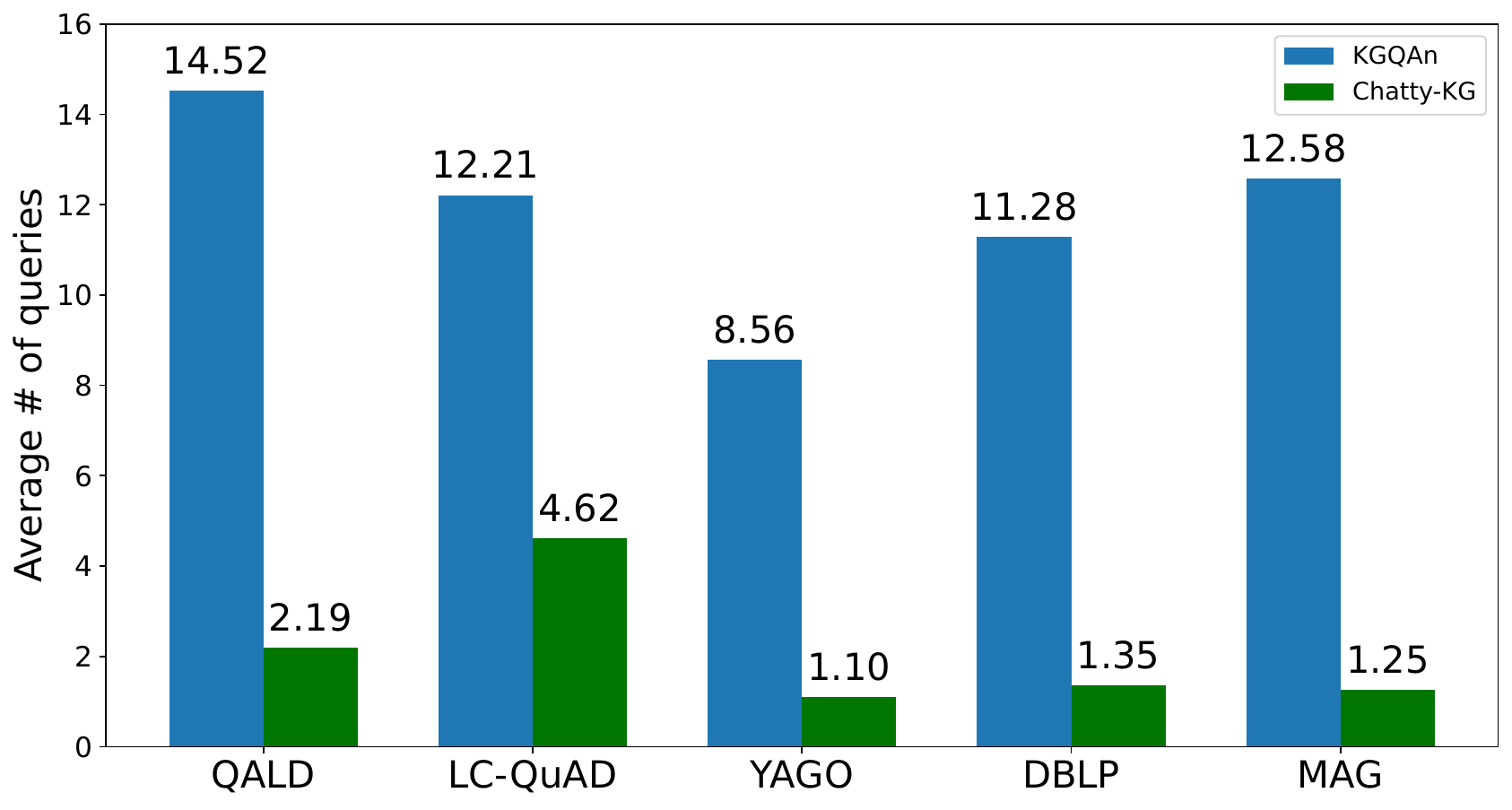}
  \vspace*{-5ex}
  \caption{
  Average number of queries per question for {\kgqan} and {\sysName}. Lower values indicate better query planning, while also {\sysName} achieves higher F1, see Table~\ref{tab:differentllms}.}
  \label{fig:numqueries}
  \vspace*{-3ex}
\end{figure}

Improved question understanding helps downstream accuracy, but gains are not always linear because errors can still happen in later stages. The remaining errors (lighter shades in \autoref{fig:question_understanding}) come from structural mismatches, entity linking, and query selection. In QALD-9, many failures stem from mismatches between natural language and KG structure. For example, \textit{``How many emperors did China have?''} should link to \textit{Emperor of China}, not \textit{China}. Likewise, \textit{``Who killed Caesar?''} should retrieve \textit{Assassins of Julius Caesar}, not just \textit{Caesar}. Some questions also rely on vague KG predicates (e.g., \textit{subject}), which do not clearly capture the intended semantics. In these cases, question parsing is correct, but the KG organizes knowledge differently than natural language.
Entity linking errors occur when the model chooses the wrong candidate due to ambiguity or similar entity names. For example, in ``\textit{Are Taiko some kind of Japanese musical instrument?}'', the system links to \url{http://dbpedia.org/resource/Japanese_musical_instrument} instead of  \url{http://dbpedia.org/class/yago/WikicatJapaneseMusicalInstruments}. 

Query selection errors arise when the agent either rejects all queries or picks one with the wrong predicate. For instance, in ``\textit{How many grand-children did Jacques Cousteau have?}'', it selects \textit{relative} instead of \textit{child}. These errors can cascade: a wrong entity link weakens query candidates, causing incorrect rejection or ranking. Chatty-KG retries up to three times to reduce such failures while controlling latency and cost. Despite these challenges, it still cuts both overall errors and QU-specific errors by a large margin.

\noindent \textbf{System Efficiency:}
We evaluate {\sysName}'s runtime efficiency against {\kgqan} based on average response time and average number of executed queries. Both are critical for interactive systems where fast and precise answers are expected. Results are shown in Figures~\ref{fig:responsetime} and~\ref{fig:numqueries}. Figure~\ref{fig:responsetime} breaks down response time into three stages: Question Understanding (QU), Linking, and Execution \& Filtration (E\&F). {\sysName} (C) consistently outperforms {\kgqan} (K) in all stages, with the largest gains in the E\&F phase. For example, on MAG, {\kgqan} takes 19.4 seconds per question, while {\sysName} completes in just 3.1 seconds. This demonstrates {\sysName}'s scalability to large graphs. Figure~\ref{fig:numqueries} shows that {\sysName} executes up to 10× fewer queries than {\kgqan}. This reduction lowers response time and improves precision by avoiding irrelevant queries, as shown in Figures~\ref{fig:responsetime} and~\ref{fig:numqueries}. 
%
%
Beyond response time, we also measured cost on QALD-9 ($150$ questions). Running {\sysName} with GPT-4o used $507$ requests, $326$K input tokens, and $7.73$K output tokens, for a total cost of $\$0.88$. This shows that {\sysName} remains cost-effective even with multiple LLM calls.\shorten


\begin{table}[t]
\centering
\caption{
Results of running {\sysName} with the translation extension module on the 7 languages of QALD-9. 
}
\label{tab:multilingual_transl}
\begin{tabular}{lccccccc}
\hline
\textbf{Lang} & \textbf{en} & \textbf{de} & \textbf{pt} & \textbf{hi} & \textbf{fr} & \textbf{nl} & \textbf{it} \\
\hline
\textbf{P} & 62.29 & 66.31 & 63.62 & 65.20 & 64.21 & 65.80 & 62.97 \\
\textbf{R} & 40.83 & 42.16 & 42.14 & 39.36 & 37.72 & 36.40 & 37.14 \\
\textbf{F1} & 49.33 & 50.79 & 50.70 & 49.08 & 47.52 & 46.87 & 46.72 \\
\hline
\end{tabular}
\end{table}

\begin{table}[t]
\vspace*{-2ex}
\centering
\footnotesize
\setlength{\tabcolsep}{4pt} 
\caption{
Human and model evaluation across quality dimensions, including inter-annotator agreement. 
}
\vspace*{-3ex}
\begin{tabular}{lccccc|ll}
\toprule
\textbf{Metric} & \textbf{ChatGPT} & \textbf{Gemini} & \textbf{H1} & \textbf{H2} & \textbf{H3} & \textbf{Avg} & \textbf{Std} \\
\midrule
\textbf{Response Quality}     & 4.48 & 4.64 & 4.50 & 4.43 & 4.22 & 4.45 & 0.15 \\
\textbf{Fluency}              & 4.92 & 4.86 & 4.88 & 4.76 & 4.86 & 4.86 & 0.06 \\
\textbf{Dialogue Coherence}   & 4.52 & 4.52 & 4.50 & 4.45 & 4.57 & 4.51 & 0.04 \\
\midrule
\multicolumn{8}{c}{\textbf{Weighted Cohen’s $\kappa$ (Human–Human): \ \ \ \ 0.53 (Moderate Agreement)}} \\
\multicolumn{8}{c}{\textbf{Weighted Cohen’s $\kappa$ (Human–LLM): \ \ \ \ \ \ \ \ \ 0.56 (Moderate Agreement)}} \\
\bottomrule
\end{tabular}
\label{tab:evaluation_scores}
\vspace*{-3ex}
\end{table}

\subsection{Analyzing {\sysName} and Discussion}

\noindent \textbf{Multilingual Support}
is enabled by adding an LLM translation module to the Chat Agent, which converts non-English questions to English before processing. This allows a single pipeline, since the KG and SPARQL entities are in English.
On QALD-9, {\sysName} yields consistent scores across 7 languages (Table~\ref{tab:multilingual_transl}), with F1 ranging from \textbf{46.72} (it) to \textbf{50.79} (de) vs.\ \textbf{49.33} (en). Some languages outperform English as translation often clarifies entity mentions, improving linking, while lower scores reflect translation quality variance. We analyzed 11 languages in total\footnote{See \href{https://github.com/CoDS-GCS/Chatty-KG/blob/main/Appendix.pdf}{\color{blue}\ul{supplementary materials}} for more details and examples.}.\shorten 

\noindent \textbf{Human and LLM Evaluation:}
To assess conversational quality, we used \textbf{five evaluators}: three human annotators and two LLMs (\textsc{ChatGPT}, \textsc{Gemini}).\footnote{Full protocol and examples are in the \href{https://github.com/CoDS-GCS/Chatty-KG/blob/main/Appendix.pdf}{\color{blue}\ul{supplementary materials.}}.} Evaluators rated each response on \textit{Response Quality} and \textit{Fluency}, and scored the full dialogue for \textit{Dialogue Coherence}. All scores used a 1–5 scale.
As shown in Table~\ref{tab:evaluation_scores}, {\sysName} achieved strong results: \textbf{4.45} in Response Quality, \textbf{4.86} in Fluency, and \textbf{4.51} in Dialogue Coherence, with low variance. Human–human agreement reached \textbf{$\kappa = 0.53$}, and human–LLM agreement \textbf{$\kappa = 0.56$}, indicating \textit{moderate agreement}. This reflects consistent judgments across annotators and reliable quality ratings.\shorten

\begin{table}[t]
 \caption{
 Number of questions solved by CHatty-KG across different question types in KGQA benchmarks. Results are shown as system solved/total questions format.
 }
 \label{tab:maintaxonomy}
 \centering
 \begin{tabular}{lcccccc}
 \cline{1-6}
  \textbf{Benchmark}
  & \textbf{Factoid}
  & \textbf{Count}
  & \textbf{Boolean}
  & \textbf{Q Len}
  & \textbf{Preds}\\
  \cline{1-6}
  QALD-9 & 63/133 & 2/13 & 0/4 & 7.52 & 1.77\\
  YAGO-B & 77/99 & - & 0/1 & 6.97 & 2.00\\
  DBLP-B & 70/92 & 7/8 & - & 9.73 & 1.72\\
  MAG-B & 58/81 & 16/17 & 2/2 & 8.63 & 1.87\\
  DBpedia-Dia & 34/64 & 0/11 & 6/19 & 7.10 & 1.00\\
  YAGO-Dia & 48/65 & 0/3 & 10/18 & 8.21 & 1.02\\
  DBLP-Dia & 48/59 & 2/4 & 17/27 & 11.62 & 1.01\\
  Wikidata-Dia & 36/61 & 1/5 & 15/25 & 9.05 & 1.03\\
  \cline{1-6}
 \end{tabular}
\end{table}

\noindent \textbf{Question Type Analysis:}
KGQA systems are best for fact-based queries (who, what, where, when, count) and are less suitable for open-ended ones that need external reasoning (why, how). We group questions into three types: \textbf{Factoid} (entity or literal answers), \textbf{Count} (requires counting), and \textbf{Boolean} (yes/no verification).\footnote{More details are in the \href{https://github.com/CoDS-GCS/Chatty-KG/blob/main/Appendix.pdf}{\color{blue}\ul{supplementary materials.}}.}
Table~\ref{tab:maintaxonomy} reports {\sysName}'s performance across these types. Factoid questions dominate benchmarks and {\sysName} solves most of them across datasets. Count and Boolean accuracy varies by dataset, as these tasks are more sensitive to query structure and KG coverage. 
We report average question length (no. of words, \textbf{Q Len}) and average predicates per query (\textbf{Preds}).
Dialogue datasets use fewer predicates since context carries across turns, reducing the need to explicitly specify all relations in each query.\shorten

\textbf{Training-Free Prompting:} Chatty-KG uses task-aware prompting rather than training. Simple agents use zero-shot prompts (e.g., Rephraser), while more complex agents use few-shot prompts (e.g., QIR) and chain-of-thought when needed. This avoids dataset creation and fine-tuning, enabling fast deployment. Although light fine-tuning could help, it requires task- and KG-specific data for each module, reducing flexibility on new or evolving graphs. Empirically, prompting alone outperforms KGQAn (which fine-tunes for question understanding) as shown in Figure~3. Prompting offers scalability, cross-KG generalization, and practical deployment without specialized training data.

\section{Conclusion}
Conversational QA over KGs enables accurate and timely access to enterprise and domain knowledge. Existing systems remain limited, mostly single-turn and dependent on heavy preprocessing or fine-tuning. General LLMs also lack grounding in structured data, reducing reliability on private or evolving graphs.
{\sysName} combines structured reasoning with LLM dialogue through a modular multi-agent design for multi-turn KGQA. It uses LLM-powered agents for context tracking, linking, and query planning without KG-specific training, enabling domain transfer, coherent dialogue, and efficient SPARQL generation across diverse graphs.
%
\textbf{\textit{Key learnings}}: (1) Modular agents allow incremental improvements without system-wide changes. (2) A shared state enables coordinated execution. (3) Agent-level validation reduces hallucination by catching errors early. (4) Bounded retries prevent infinite loops when tasks exceed LLM capability. (5) The multi-agent design adds little latency and debugging becomes more complex. These choices deliver accuracy, adaptability, and practical performance for real-world KGQA.\shorten


\bibliographystyle{ACM-Reference-Format}
\bibliography{references}
\end{document}